\documentclass[twocolumn,pagebackref,breaklinks,colorlinks,allcolors=iccvblue]{fairmeta}

\usepackage{amsmath} 
\usepackage{algorithmic}
\usepackage{array} 
\usepackage{subcaption}
\usepackage{dblfloatfix}
\usepackage{url} 

\usepackage{microtype}
\usepackage[utf8]{inputenc} 
\usepackage[T1]{fontenc}    
\usepackage{booktabs}       
\usepackage{siunitx}

\usepackage{enumitem}
\setlist[enumerate]{noitemsep, topsep=0pt}
\setlist[itemize]{noitemsep, topsep=0pt}

\usepackage{marvosym} 
\usepackage{bm} 
\usepackage{bbm} 
\usepackage{amssymb}   
\usepackage{mathrsfs} 
\usepackage{tabularray} 
\UseTblrLibrary{booktabs}       
\usepackage{tabularx} 
\usepackage[flushleft]{threeparttable} 
\usepackage{colortbl} 
\usepackage{makecell} 
\usepackage{multirow} 
\usepackage{nicefrac} 
\usepackage{soul} 
\setul{0.5ex}{0.3ex}

\usepackage{cellspace}

\usepackage{xspace} 

\usepackage{pifont} 
\newcommand{\xmark}{\ding{53}}%

\definecolor{Gray}{rgb}{0.9,0.9,0.9}
\definecolor{LightCyan}{rgb}{0.88,1,1}
\newcolumntype{a}{>{\columncolor{Gray}}c}
\newcolumntype{b}{>{\columncolor{white}}c}
\definecolor{iccvblue}{rgb}{0.21,0.49,0.74}

\usepackage{url}
\usepackage{makecell}

\usepackage{epsfig}
\usepackage{graphicx}
\usepackage{amsmath}
\usepackage{amssymb}
\usepackage{fontawesome5}
\usepackage{graphicx}
\usepackage{amsmath}
\usepackage{amssymb}
\usepackage{booktabs}
\usepackage[ruled,vlined]{algorithm2e}
\usepackage{algorithmic}
\usepackage{colortbl} 
\usepackage{booktabs} 
\usepackage{multirow, booktabs, graphicx, xcolor, colortbl, arydshln, tikz}
\usetikzlibrary{tikzmark}
\usepackage{makecell}
\usepackage{siunitx} 
\usepackage{mdframed}
\usepackage{tablefootnote} 

\newcolumntype{g}{>{\columncolor{gray!30}}c}

\usepackage{tcolorbox}
\tcbuselibrary{most}

\definecolor{citecolor}{HTML}{0071bc}
\hypersetup{
    colorlinks=true,%
    citecolor=citecolor,%
    filecolor=citecolor,%
    linkcolor=citecolor,%
    urlcolor=citecolor
}

\tcbuselibrary{skins}

\definecolor{userbg}{RGB}{245, 245, 245}
\definecolor{userborder}{RGB}{210, 229, 255}
\definecolor{userfont}{RGB}{0, 0, 0}

\tcbset{
  enhanced,
  boxrule=1pt,
  colback=userbg,
  colframe=userborder,
  fonttitle=\bfseries,
  coltitle=userfont,
  boxsep=5pt,
  arc=5pt,
  outer arc=5pt,
  left=10pt,
  right=10pt,
  top=2pt,
  bottom=2pt,
  attach boxed title to top left={yshift=-0.25\baselineskip-1mm,xshift=2mm},
  boxed title style={
    colback=userborder,
    sharp corners,
    rounded corners=northeast,
    arc=3pt,
    outer arc=3pt,
    boxrule=0pt,
    boxsep=2pt,
  },
}
\usepackage{pgf}
\usepackage{adjustbox}

\usepackage{multirow}
\usepackage{colortbl} 
\usepackage{booktabs}
\usepackage{tcolorbox} 
\usepackage{enumitem}
\usepackage{graphicx}
\usepackage{multirow}
\usepackage{array}
\definecolor{listcolor}{RGB}{50,120,230}
\usepackage{arydshln}
\usepackage{url}
\usepackage{caption}

\newcounter{researchquestion}

\newcommand{\researchquestion}[2][]{%
  \vspace{0.8em}
  \refstepcounter{researchquestion}
  \begin{tcolorbox}[
    enhanced,
    colback=blue!5,
    colframe=blue!70!black,
    fonttitle={\fontsize{10.5pt}{12.8pt}\selectfont\bfseries\color{blue!20!black}},  
    title=Question \theresearchquestion,
    toprule=1.5pt,
    bottomrule=0.8pt,
    leftrule=0.8pt,
    rightrule=0.8pt,
    left=6pt,
    right=6pt,
    top=6pt,
    bottom=6pt,
    boxsep=3pt
  ]
  \normalsize #2
  \end{tcolorbox}
  \ifx\\#1\\\else\label{rq:#1}\fi
  \vspace{0.5em}
}

\title{DenoDet V2: Phase-Amplitude Cross Denoising for SAR Object Detection}

\author[1,*]{Kang Ni}
\author[2,*]{Minrui Zou}
\author[2]{Yuxuan Li}
\author[2,3]{Xiang Li}
\author[4]{Kehua Guo}
\author[2,3]{Ming-Ming Cheng}
\author[2,3\dagger]{Yimian Dai}

\affiliation[1]{School of Computer Science, Nanjing University of Posts and Telecommunications}
\affiliation[2]{VCIP Lab, Computer Science, NKU}
\affiliation[3]{NKIARI, Futian, Shengzhen}
\affiliation[4]{School of Computer Science, Central South University}
\contribution[*]{Equal Contribution}
\contribution[\dagger]{Corresponding Author}

\abstract{
One of the primary challenges in Synthetic Aperture Radar (SAR) object detection lies in the pervasive influence of coherent noise.
As a common practice, most existing methods, whether handcrafted approaches or deep learning-based methods, employ the analysis or enhancement of object spatial-domain characteristics to achieve implicit denoising.
In this paper, we propose DenoDet V2, which explores a completely novel and different perspective to deconstruct and modulate the features in the transform domain via a carefully designed attention architecture.
Compared to DenoDet V1, DenoDet V2 is a major advancement that exploits the complementary nature of amplitude and phase information through a band-wise mutual modulation mechanism, which enables a reciprocal enhancement between phase and amplitude spectra.
Extensive experiments on various SAR datasets demonstrate the state-of-the-art performance of DenoDet V2. Notably, DenoDet V2 achieves a significant 0.8\% improvement on SARDet-100K dataset compared to DenoDet V1, while reducing the model complexity by half. The code is available at \url{https://github.com/GrokCV/GrokSAR}.

}

\date{Augest 12, 2025}

\project{\href{https://github.com/GrokCV/GrokSAR}{\sffamily \fontsize{8.8pt}{11pt}\selectfont \texttt{https://github.com/GrokCV/GrokSAR}}}

\begin{document}
\maketitle


\section{Introduction} \label{sec:introduction}


Synthetic Aperture Radar (SAR) has revolutionized the field of remote sensing, offering an unrivaled capability to capture high-resolution imagery regardless of lighting conditions or weather obscurity.
Despite the progress made by deep learning methods~\cite{JSTARS2022Morphological,zhang2025rsar,sm3det}, most of them are straightforward adaptations from generic object detection in computer vision, without fully considering the unique characteristics of SAR data and its associated challenges. As SAR is a coherent imaging system, its images intrinsically contain \textbf{unavoidable speckle noise}, overlaid on the objects, significantly increasing the difficulty of object detection and identification.

\begin{figure}[!t]
  \centering
  \includegraphics[width=0.45\textwidth]{"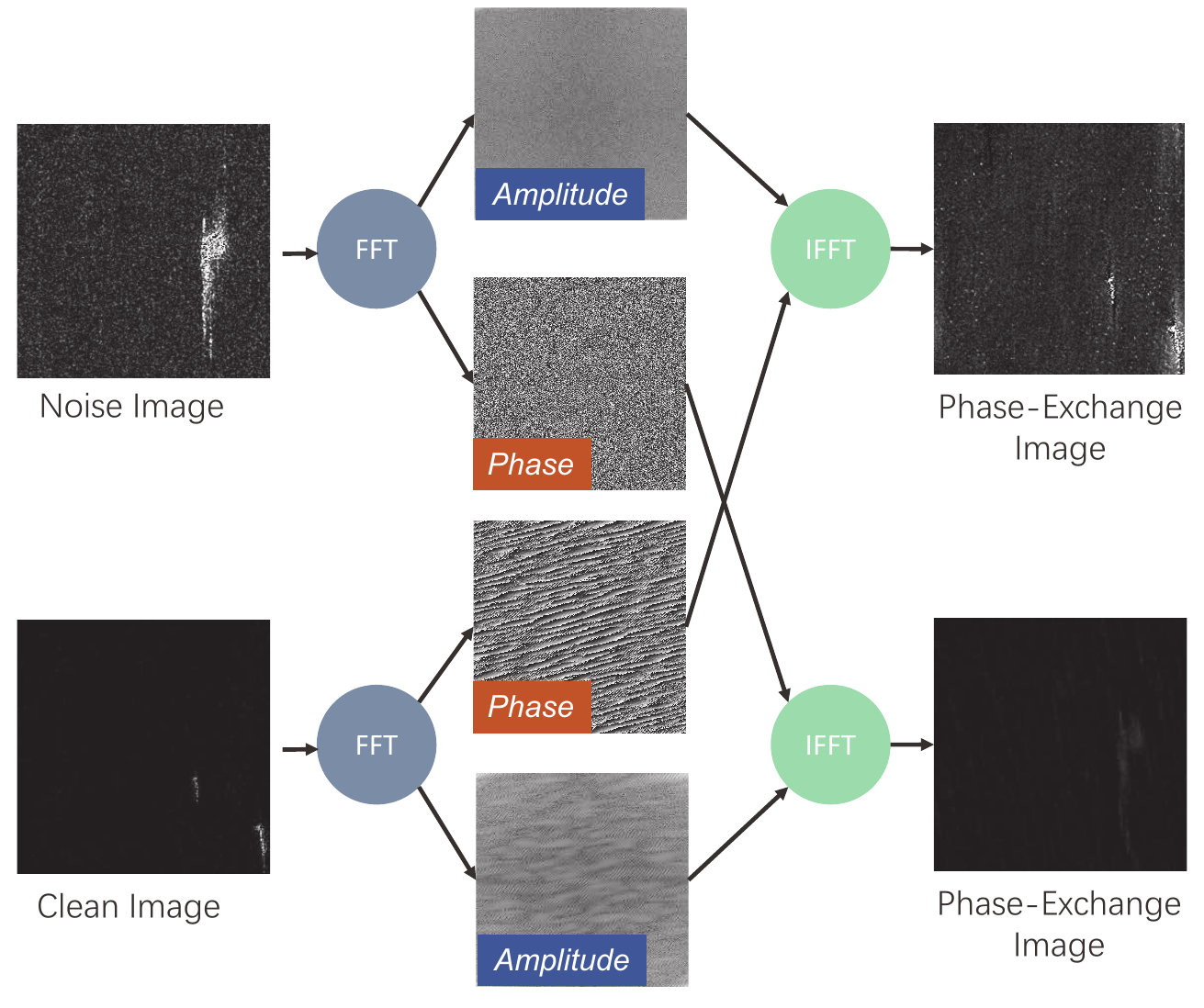"}
    \vspace{-.5\baselineskip}
  \caption{An example that the phase spectrum is more robust to noise. In the images after phase information exchange, the object contours from the noise images are clearly transferred while introducing only a small amount of noise interference.}
  \vspace{-1.\baselineskip}
  \label{fig:phase_vs_amplitude}
\end{figure}




To tackle the aforementioned challenges, recent works have integrated frequency-domain processing within deep learning frameworks, aiming to harness its potential in separating the speckle noise.
Zhao \textit{et al.} combined a morphological network with a feature pyramid fusion structure for speckle noise suppression \cite{JSTARS2022Morphological}.
Similarly, Wang \textit{et al.} employed a dual-backbone structure integrated with Haar wavelet transform \cite{TGRS2023SpatialFrequency}, adept at capturing both global and intricate textural details of maritime objects.
Complementing these, Li \textit{et al.} unveiled a network synergizing a feature pyramid network with a polar Fourier transform \cite{TGRS2022Multidimensional}, procuring rotation-invariant features and delivering enhanced performance in ship object detection.

Although these methods have demonstrated improved performance in SAR object detection by exploiting frequency-domain information, they still have some limitations.
\begin{enumerate}
    \item \textbf{Isolated Processing of Amplitude and Phase:} These approaches tend to process amplitude and phase information in isolation. Besides, these methods predominantly focus on the amplitude components, neglecting the phase details which are crucial for preserving the original spatial relationships and structural integrity of the objects as shown in Fig. \ref{fig:phase_vs_amplitude}.
    \item \textbf{Increased Model Complexity and Over-Interaction:} These approaches involve interactions across the entire spectrum, which not only escalates computational complexity but may also induce an overprocessing of frequency-domain features. This overprocessing can manifest as either an excessive smoothing of object features, which diminishes detection accuracy.
\end{enumerate}




Recent studies~\cite{chen2021amplitude} have highlighted the pivotal role of phase information in maintaining the structural integrity and spatial relationships within convolutional neural networks, crucial for robust object recognition.
This contrasts with amplitude information, which, though important, is more vulnerable to noise and perturbations. This insight is particularly salient in SAR imagery, where coherent speckle noise severely hampers object detection.
Here, phase information provides a resilient feature, remaining stable amidst noise that often distorts amplitude, thereby preserving critical scene details.
\textit{It raises a natural question: Can phase information be leveraged at the feature level in the frequency domain to mitigate the effects of coherent speckle noise on the amplitude spectrum, and vice versa?}

Motivated by this intriguing possibility, in this paper, we extend the idea of \textbf{attention as feature denoising} by proposing DenoDet V2, a novel method that leverages the complementary nature of amplitude and phase information through a mutual guidance mechanism for SAR image object detection.
Central to our approach is a phase-guided soft-thresholding mechanism that exploits the robustness of phase information to adaptively filter noise from the amplitude spectrum. 
DenoDet V2 not only enhances the signal-to-noise ratio of the object features but also uses the refined amplitude spectrum to further improve phase accuracy, ensuring the preservation of essential object details.
\textit{To our knowledge, DenoDet V2 is the first to implement such a reciprocal, reference-based feature denoising strategy.}

Specifically, DenoDet V2 involves a strategic interplay within the Key and Value component of the self-attention~\cite{vaswani2017attention} module, where the roles of phase and amplitude are dynamically interchanged.
To further tailor this process to the intricate nature of SAR imagery, we partition the frequency spectrum into a grid of $N \times N$ local bands, within which phase and amplitude interact exclusively. This localized, band-specific approach ensures fine-grained feature denoising that is acutely attuned to each band's unique characteristics, meticulously preserving vital spatial and structural information. 

In summary, we advance the state-of-the-art in SAR image object detection with our contributions as follows:
\begin{enumerate}
    \item 
    We propose a novel concept, \textbf{\textit{Attention as Phase-Amplitude Cross Denoising}}, which leverages the inherent stability of phase information to guide the feature denoising of amplitude in SAR imagery. 
    \item 
    We propose the Phase-Amplitude Token Exchange (PATE) module as a core component of our DenoDet V2. The PATE module implements a dual guidance mechanism, where phase and amplitude information mutually enhance each other.
    \item 
    Our DenoDet V2 achieves the state-of-the-art performance
    across various SAR datasets, especially \textbf{ranking No. 1 on the largest benchmark SARDet-100K\footnote{\url{https://github.com/GrokCV/GrokSAR}}}.
\end{enumerate}

\section{Related Work} \label{sec:related}

\subsection{SAR Image object Detection}


Recent years have witnessed significant advancements in the application of deep learning techniques to Synthetic Aperture Radar (SAR) object detection.
For instance, Sun et al. \cite{sun2021oriented} introduced a method that utilizes strong scattering points to enhance SAR object detection, particularly effective in near-shore scenes where land interference is prevalent. 
Similarly, Ke et al. \cite{ke2021sar} improved the adaptability of Faster R-CNN for SAR images by integrating deformable convolution kernels, which better accommodate the geometric variations of ships in SAR imagery. 
Additionally, PVT-SAR \cite{zhou2022pvt} leveraged Pyramid Vision Transformers (PVT) to extract multi-scale features through a self-attention mechanism, demonstrating substantial gains in detection performance.

Despite these advancements, the inherent issue of speckle noise in SAR images remains a formidable challenge. 
This noise, characteristic of the radar imaging process, can obscure critical object details and lead to high false alarm rates. 
Recent studies have explored the application of deep learning to speckle denoising in SAR images.
For instance, 
Shen \textit{et al.} \cite{shen2021sar} proposed a recursive deep CNN prior model that decouples the data-fitting and regularization terms for improved denoising.

Nevertheless, these studies generally \textit{treat denoising and object detection as separate processes}.
The lack of integration makes it challenging to dynamically adjust the denoising process based on the object features, which is crucial for preserving the discriminative information for detection. 
Moreover, these denoising methods primarily focus on enhancing the spatial domain features, without fully exploiting the potential of frequency domain information.
On the contrary, our DenoDet V2 integrates a frequency-domain denoising module into a detection framework, which differs from existing methods in the paradigm of dynamic feature denoising and phase-amplitude token exchange.

\subsection{Frequency-Domain Feature Refinement}
Recent studies have explored the potential of incorporating frequency-domain information into SAR object detection to enhance performance. 
Zhao et al. \cite{zhao2018cascade} explored a visual attention mechanism that incorporates frequency-domain elements to refine ship detection in SAR images. 
Similarly, Li et al. \cite{li2022novel} developed a multidimensional domain network that leverages both spatial and frequency-domain features to detect ships in SAR imagery effectively. 
Xu et al. \cite{xu2020learning} further emphasized the importance of frequency-domain learning, proposing a method that selectively processes frequency components to optimize network input and improve computational efficiency.


Despite these advancements, current approaches  typically neglect the phase spectrum or treat phase and amplitude information in isolation, which often results in the inability to dynamically interact between phase and amplitude spectra. However, this is crucial for enhancing the detection performance in noisy SAR environments, which has been confirmed by many studies. For instance,
Chen et al. \cite{chen2021amplitude} found that CNNs heavily depend on the amplitude spectrum of images and are easily disturbed by noise. 
They proposed a data augmentation method involving phase-amplitude spectrum sample replacement to improve the generalization performance, further highlighting the significance of phase information.
In contrast to the aforementioned works, our proposed DenoDet V2 model differs in two key aspects: Phase-Amplitude Modulation and Frequency-Domain Band Division.

\section{Method} \label{sec:method}

\subsection{Overall Architecture}
\begin{figure*}[htbp]
  \centering
  \includegraphics[width=\textwidth]{"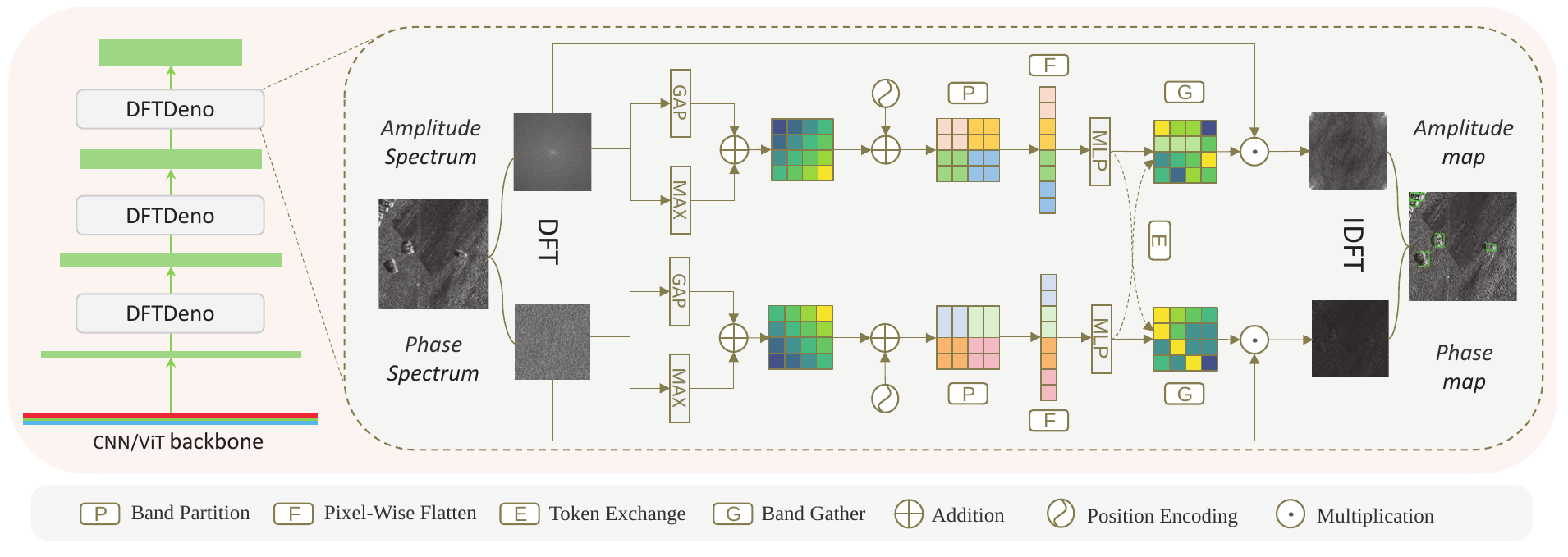"}
  \captionsetup{justification=centering}
  \caption{The overall architecture of DenoDet V2.}
  \label{fig:DenoDetV2}
  \vspace*{-1\baselineskip}
\end{figure*}

The overall architecture of our proposed DenoDet V2 is shown in the Fig. \ref{fig:DenoDetV2}. It is based on a generic object detection network and our DFTDeno module is integrated for enhancement as a plug-and-play module. Through a designed attention mechanism DFTDeno performs dynamic soft threshold denoising in the transform domain of the feature maps. In DenoDetV2, we place it in the backbone network stage to refine the process of feature map extraction.

DFTDeno module consists of a forward 2D Discrete Fourier Transform (DFT), dynamic threshold denoising, followed by an inverse 2D DFT. For a feature map $\mathbf{M}\in{\mathbb{R}^{C \times H \times W}}$, the 2D DFT forward function can be defined as:
\begin{equation}
 \mathbf{m}_{c,u,v} = \sum_{h=0}^{H-1}\sum_{w=0}^{W-1}\mathbf{M}_{c,h,w}e^{ - 2{\pi}i(\frac{{uh}}{H} + \frac{{vw}}{W})}.
\end{equation}
According to Euler's formula, it can also be expressed as:
\begin{equation}
\scalebox{0.8}{
$\displaystyle \sum_{h=0}^{H-1}\sum_{w=0}^{W-1}\mathbf{M}_{c,h,w}\left [cos2{\pi}\left(\frac{{uh}}{H} + \frac{{vw}}{W}\right) - isin2{\pi}\left(\frac{{uh}}{H} + \frac{{vw}}{W}\right)\right]$,
}
\label {formulation: DFT}
\end{equation} where $H$ and $W$ denotes the height and width of features, a pair of $h, u \in [0, H - 1]$, and $w, v \in [0, W - 1]$ represents the coordinate position of statistics, and $\mathbf{m}_{c,u,v}$ means the value located at coordinates $(u, v)$ in channel $c$ of the tensor after the forward DFT.
Via Eq.\ (\ref{formulation: DFT}), 
we can decompose the signal component after DFT into real and imaginary part:
\begin{gather}
\mathcal{R}_{c, u, v} = \sum_{h=0}^{H-1}\sum_{w=0}^{W-1}\mathbf{M}_{c,h,w}cos2{\pi}\left(\frac{{uh}}{H} + \frac{{vw}}{W}\right),\\
\mathcal{I}_{c, u, v} = -\sum_{h=0}^{H-1}\sum_{w=0}^{W-1}\mathbf{M}_{c,h,w}isin2{\pi}\left(\frac{{uh}}{H} + \frac{{vw}}{W}\right).
\end{gather}

Next, the real part $\mathcal{R}\in{\mathbb{R}^{C \times H \times W}}$ and imaginary part $\mathcal{I}\in{\mathbb{R}^{C \times H \times W}}$ undergo two non-linear transformations to extract the amplitude spectrum $\mathcal{A}$ and phase spectrum $\mathcal{P} \in [-\pi,\pi)$ of the frequency-domain signal as follows:
\begin{gather}
\mathcal{A}_{c, u, v} = \sqrt{\mathcal{R}^2_{c, u, v} + \mathcal{I}^2_{c, u, v}},\\
\mathcal{P}_{c, u, v} = arctan2\left(\frac{{\mathcal{I}_{c, u, v}}}{\mathcal{R}_{c, u, v}}\right).
\end{gather}

Eq.\ (\ref{formulation: DFT}) reveals that each signal component is derived from pixel-level values in the original feature map, inherently encoding partial global information. To process such highly aggregated information, we employ an efficient attention mechanism for signal modulation. This operation suppresses noise while enhancing informative features through the following transformation:
\begin{gather}
\mathcal{\hat{A}} = \mathbf{G}(\mathcal{A, P}) \odot \mathcal{A}, \quad \mathcal{\hat{P}} = \mathbf{G}(\mathcal{A, P}) \odot \mathcal{P},
\end{gather}
where $\mathcal{\hat{A}}$ and $\mathcal{\hat{P}}$ represent the signal's amplitude and phase after modulation, $\odot$ means element-wise multiplication and $\mathbf{G(\cdot)}\in \mathbb{R}^{C \times H \times W}$ is the output attention map of designed module $\mathbf{G}$, which will be discussed in section \ref{sec:token exchange} in detail.

Finally, we recombine the modulated amplitude and phase information and restore frequency signals to the spatial domain via the Inverse Discrete Fourier Transform (IDFT) as follows:
\begin{gather}
 \mathbf{\hat{m}}_{c,u,v} = \mathcal{\hat{A}}_{c,u,v} \cdot (cos\mathcal{\hat{P}}_{c,u,v} + isin\mathcal{\hat{P}}_{c,u,v}),
\label{formulation: recombine} \\
 \mathbf{\hat{M}_{c,h,w}} = \sum_{u=0}^{H-1}\sum_{v=0}^{W-1}\mathbf{\hat{m}}_{c,u,v}e^{ 2{\pi}i(\frac{{uh}}{H} + \frac{{vw}}{W})}.
\end{gather}
For brevity, the normalization coefficients in both the forward and inverse transforms are omitted.

In the final implementation, instead of original phase features, we employed $cos(\mathcal{\hat{P}})$ and $sin(\mathcal{\hat{P}})$ to decouple the phase information according to Eq.\ (\ref{formulation: recombine}), effectively mitigating the angular boundary discontinuity issue. And the trigonometric identity was subsequently applied to harmonize these two orthogonal components, thereby ensuring parity in the resultant phase representation through rigorous mathematical equivalence.

\subsection{Band-wise Partition Self-Attention}
In the frequency domain, adjacent features do not necessarily exhibit local correlations.
Therefore, the long-context modeling capability of the self-attention mechanism becomes critically important. Additionally, in the transform domain, each sampling point corresponds to a specific frequency component.
Therefore, we have adopted the self-attention mechanism as the foundational architecture for band-wise signal modulation. Let $\mathbf{X}\in{\mathbb{R}^{C \times H \times W}}$ denotes the frequency domain tensor after DFT, the output of self-attention block for signal modulation is calculated via:
\begin{gather}
 \mathbf{\hat{X}} = \mathbf{P}_{\operatorname{Max}}(\mathbf{X})+ \mathbf{P}_{\operatorname{Avg}}(\mathbf{X}),\\
 \mathbf{\widetilde{S}} = BPSA(\mathbf{\hat{X}}), \quad \mathbf{\widetilde{X}} = \mathbf{\widetilde{S}} \odot \mathbf{X},
\end{gather}
where $\mathbf{\hat{X}} \in \mathbb{R}^{H*W}$ is the frequency spectrum attention map, $\mathbf{P}_{\operatorname{Max}}$ and $\mathbf{P}_{\operatorname{Avg}}$ means max and average pooling operation among channel dimension, and BPSA denotes the band-wise partition self-attention, $\mathbf{\widetilde{X}} \in \mathbb{R}^{C \times H \times W}$ means the output signal after modulation. In basic band-wise self-attention (BSA), the input feature $\mathbf{\hat{X}}$ is split by band frequency as $\{ \mathbf{\hat{X}}_{1},\mathbf{\hat{X}}_{2},...,\mathbf{\hat{X}}_{d} \}$ and $d=H*W$ for all of coordinate pixel represents an identity frequency signal in basic BSA. The result of each band from BSA can be produced as:
\begin{gather}
 \mathbf{Q}_{i}=\mathbf{\hat{X}}_{i}\mathbf{W}^{q}_{i},\quad \mathbf{K}_{i}=\mathbf{\hat{X}}_{i}\mathbf{W}^{k}_{i},\quad \mathbf{V}_{i}=\mathbf{\hat{X}}_{i}\mathbf{W}^{v}_{i},
\label {formulation: QKV}\\
 \mathbf{\widetilde{S}}_{i} = SoftMax(\mathbf{Q}_{i}{\mathbf{K}}_{i}^{T}/ \sqrt{d} + \mathbf{E})\mathbf{V}_{i},
\label {formulation: attention} \\
 \mathbf{\widetilde{S}} = MLP(\sum_{i=0}^{d}\mathbf{\widetilde{S}}_{i}),
\end{gather}
where $\mathbf{W}^{q}_{i}, \mathbf{W}^{k}_{i}, \mathbf{W}^{v}_{i} \in \mathbb{R}^{d}$ are learnable weight for producing query $\mathbf{Q}_{i} \in \mathbb{R}^{H*W}$, keys $\mathbf{K}_{i} \in \mathbb{R}^{H*W}$, and value $\mathbf{V}_{i} \in \mathbb{R}^{H*W}$. $\mathbf{E}$ denotes the positional encoding parameters, $\sqrt{d}$ is a reweigh factor relates with the number of frequency band. Moreover, within the feature map after DFT transformation, the frequencies exhibit a non-monotonic variation with their 2D coordinates, initially decreasing before rising as the coordinates extend. Specifically, the signal frequencies are centrally symmetric about the point $(\frac{2}{H}, \frac{2}{W})$. To facilitate subsequent modeling, we applied a central shift to this relationship, The visualization of this special signal characteristic is shown in Fig. \ref{fig:dft} in the supplementary material.

\begin{figure}[h]
  \centering
  \includegraphics[width=.35\textwidth]{"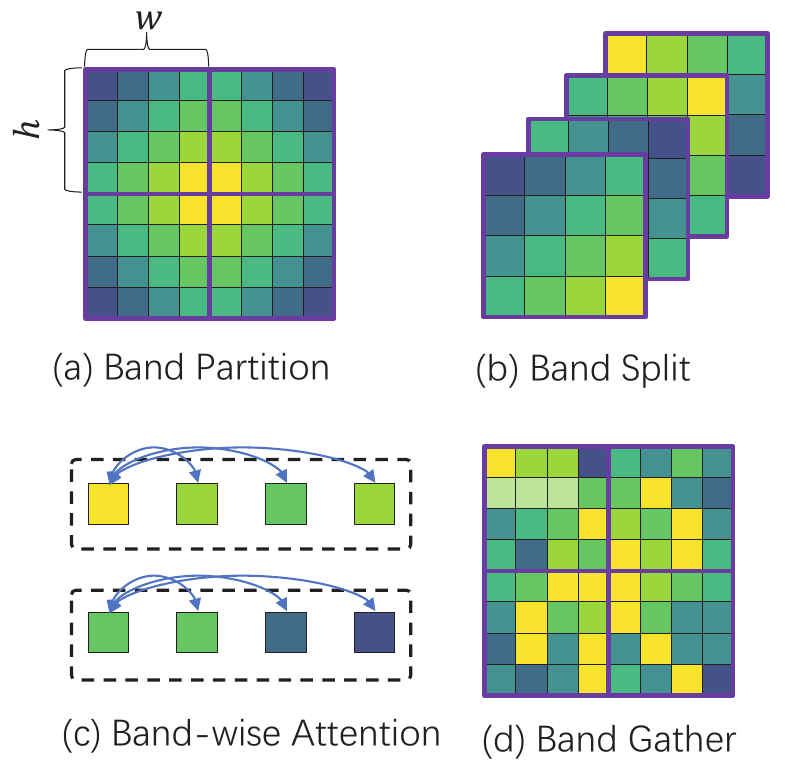"}
  \vspace{-.5\baselineskip}
  \caption{Process of band-wise partition attention module. All frequency of amplitude and phase signals are partitioned by ($h$,$w$) kernel size unfolding operation, then the attention is only performed on signals within the same group, after that, all of bands are gathered by folding operation into spatial dimension.}
  \label{fig:partition}
  \vspace*{-1\baselineskip}
\end{figure}

Based on this spectral property, we propose the band-wise partition self-Attention mechanism strategically decomposes the global attention operation into parallelizable sub-band computations as visualized in Fig.\ref{fig:partition}, $h$ and $w$ denotes the Vertical and horizontal stride to partition frequency bands, and in Eq. (\ref{formulation: attention}), the number of frequency band group $d$ should be rewritten as $d=\frac{H*W}{h*w}$, and $\mathbf{Q}_{i},\mathbf{K}_{i},\mathbf{V}_{i} \in \mathbb{R}^{h*w}$ denote the query, keys and value for an identity group band. In this case, our band-wise partitioning paradigm achieves dual objectives: 1) \textit{Preserving frequency-domain data distributions,} 2) \textit{Enabling regional attention with reduced dimensionality.}

\subsection{Phase and Amplitude Token Exchange} \label{sec:token exchange}

\begin{figure}[!t]
  \centering
  \includegraphics[width=.5\textwidth]{"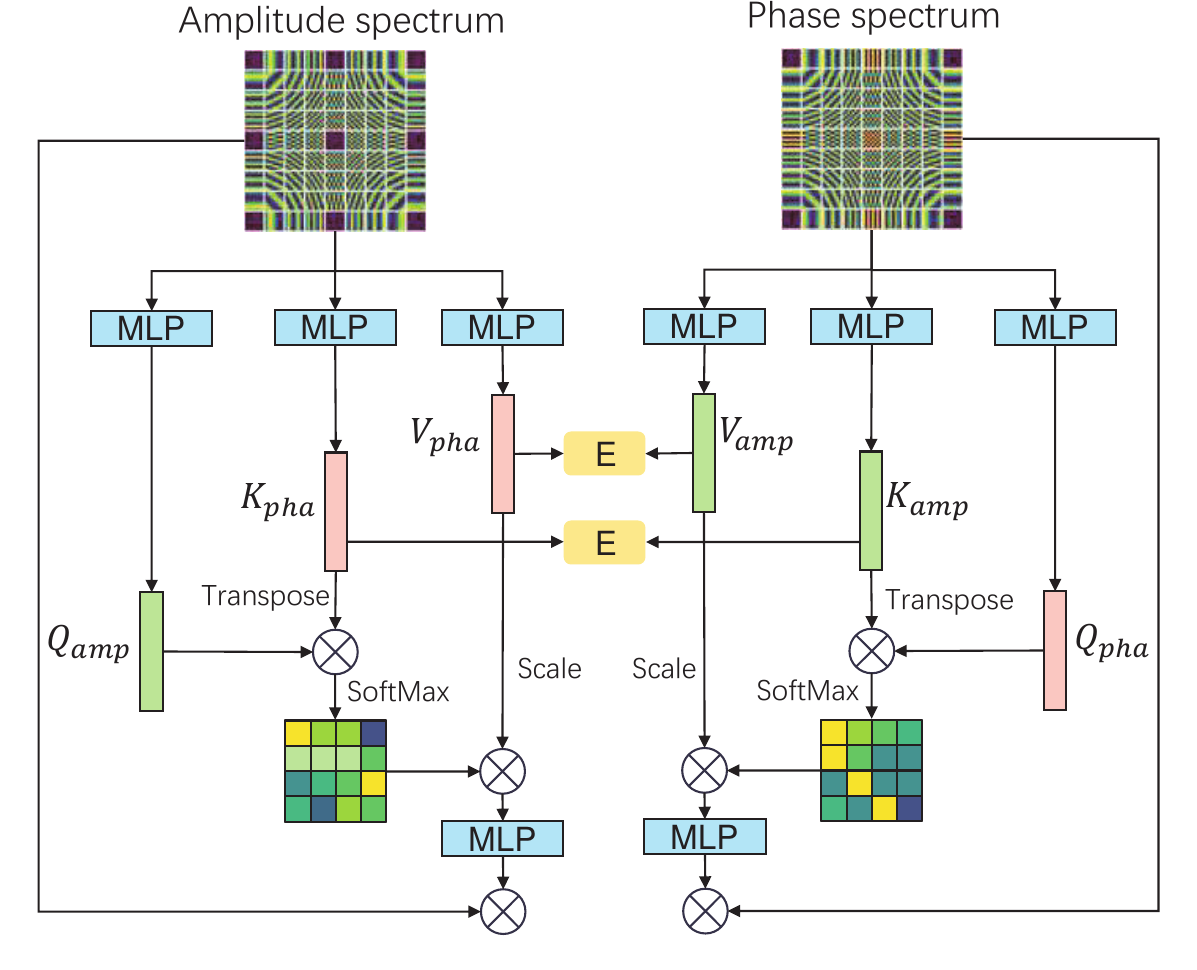"}
  \vspace{-.5\baselineskip}
  \caption{Illustration of DenoDet V2's token exchange scheme.
  }
  \vspace{-2.\baselineskip}
  \label{fig:exchange}
\end{figure}

Conventional frequency-domain denoising approaches typically implement independent processing pipelines for amplitude and phase spectral components, which reduces the model's ability to adaptively interact with the two modalities of information. To overcome this fundamental limitation, we propose a token change method into cross-spectral attention module, which allows the network to adaptively choose to use phase data to guide amplitude data, or vice versa. The process of amplitude and phase map token exchange is illustrated in Fig.~\ref{fig:exchange}.

Firstly, we employ the same stride parameter to segment the amplitude spectrum and phase spectrum into discrete groups, ensuring congruent frequency alignment between the dual modal signals within corresponding groups. These partitioned frequency components are then separately processed through MLP layers to generate modality-specific query, key, and value representations. Subsequently, we exchange the frequency tokens within the same group. This methodology establishes distinct mapping patterns within individual groups while maintaining strict inter-group information isolation, thereby preserving the integrity of intra-group frequency characteristics throughout the transformation process. For amplitude attention map $\mathcal{A}\in{\mathbb{R}^{H \times W}}$ and phase attention map $\mathcal{P}\in{\mathbb{R}^{H \times W}}$ , they are firstly partitioned as $\{ \mathcal{\hat{A}}_{1},\mathcal{\hat{A}}_{2},...,\mathcal{\hat{A}}_{d} \}$, $\{ \mathcal{\hat{P}}_{1},\mathcal{\hat{P}}_{2},...,\mathcal{\hat{P}}_{d} \}$, where $\mathcal{\hat{A}}_{i},\mathcal{\hat{P}}_{i} \in \mathbb{R}^{h \times w}$, and then, the process of token exchange within group can be defined by the following equations:

\begin{gather}
\mathbf{Q}^{\mathcal{A}}_{i}=\mathcal{\hat{A}}_{i}\mathbf{W}^{\mathcal{A}^{q}}_{i},
\mathbf{K}^{\mathcal{A}}_{i}=\mathcal{\hat{P}}_{i}\mathbf{W}^{\mathcal{P}^{k}}_{i},
\mathbf{V}^{\mathcal{A}}_{i}=\mathcal{\hat{P}}_{i}\mathbf{W}^{\mathcal{P}^{v}}_{i},\\
\mathbf{Q}^{\mathcal{P}}_{i}=\mathcal{\hat{P}}_{i}\mathbf{W}^{\mathcal{P}^{q}}_{i},
\mathbf{K}^{\mathcal{P}}_{i}=\mathcal{\hat{A}}_{i}\mathbf{W}^{\mathcal{A}^{k}}_{i},
\mathbf{V}^{\mathcal{P}}_{i}=\mathcal{\hat{A}}_{i}\mathbf{W}^{\mathcal{A}^{v}}_{i},
\end{gather}

where $\mathbf{W}^\mathcal{A}_{i}, \mathbf{W}^\mathcal{P}_{i} \in \mathbb{R}^{d}$ denote the linear mapping weight for amplitude features $\mathcal{A}$ and phase features $\mathcal{P}$. $Q_{i},K_{i},V_{i} \in \mathbb{R}^{h*w \times d}$ are queries, keys and values for each group of corresponding spectrum. In this case, the swap of tokens between amplitude and phase forces the model to interact with the two modalities of information. 

\section{Experiments} \label{sec:experiment}





\subsection{Implementation Details} \label{subsubsec:implementation}
Three benchmark datasets were evaluated under unified configurations: \textbf{SARDet-100K}~\cite{NeurIPS2024SARDet}, \textbf{SAR-Aircraft-1.0}~\cite{xian2019air}, and \textbf{AIR-SARShip-1.0}~\cite{zhirui2023sar}. All experiments employed the MMDetection framework with four RTX 4090 GPUs. Training stability was ensured through gradient clipping and synchronized batch normalization in distributed training environments. Key configurations are detailed as follows.

\textbf{SARDet-100K}: This multi-class dataset contains 116,598 images (8:1:1 split) from 10 international sub-collections, covering six maritime/land objects. Images were resized to $512 \times 512$ pixels with 50\% horizontal flipping. Training used DAdaptAdam optimizer for 12 epochs (batch size 16, LR=1.0, weight decay=0.05). 

\textbf{SAR-Aircraft-1.0}: Featuring 7 aircraft categories (3,489 training/879 test images), the dataset was processed into $512 \times 512$ sub-images with 200-pixel overlaps. Identical optimizer parameters as SARDet-100K were applied for 12 epochs, with FLOPs computed equivalently.

\textbf{AIR-SARShip-1.0}: Comprising 31 Gaofen-3 scenes (3000$\times$3000 pixels), this ship dataset generated $512 \times 512$ chips (200-pixel overlap) containing 461 annotated vessels. 





\subsection{Comparison with State-of-the-Arts} \label{subsec:sota}
\begin{table*}[h]
  \renewcommand\arraystretch{1.2}
  \centering
  \captionsetup{justification=centering}
  \caption{Comparison with SOTA methods on \textbf{SARDet-100K}.}
  \label{tab:sardet}
  \vspace{-2pt}
  \small
  \setlength{\tabcolsep}{1pt}{
  \begin{tabular}{l|ccacc}
  Method & {\textbf{FLOPs $\downarrow$}} & \small{\textbf{\#P $\downarrow$}} & \small{\textbf{mAP$\uparrow$}} & 
  \small$\mathbf{AP}$\scriptsize$_{\mathbf{S}}$\small$\uparrow$ &
  \small$\mathbf{AP}$\scriptsize$_{\mathbf{M}}$\small$\uparrow$ \\
  
  \Xhline{1pt}
    \multicolumn{5}{l}{\textit{Two-stage}}  \\
  
  \hline
  Faster R-CNN~\cite{ren2015faster}                  & 63.2G  & 41.37M                       & 39.22        & 32.55 & 47.23                 \\

  Cascade R-CNN~\cite{cai2019cascade}                 & 90.99G     & 69.17M                    & 53.55         & 49.09 & 62.89               \\

  Dynamic R-CNN~\cite{zhang2020dynamic}               & 63.2G      & 41.37M                   & 49.75        & 43.12 & 59.72 \\

  Grid R-CNN~\cite{lu2019grid}                  & 180G            & 64.47M             & 50.05         & 42.43 & 62.01 \\
  
  Libra R-CNN~\cite{pang2019libra}                   & 64.02G       & 41.64M                  & 52.09         & 45.85 & 63.52 \\

  ConvNeXt~\cite{CVPR2022ConvNeXt}                  & 63.84G          & 45.07M              & 53.15         & 45.67 & 64.55\\

  ConvNeXtV2~\cite{CVPR2023ConvNeXtV2}                 & 120G         & 110.0G                & 53.91     & 47.63 & 64.67 \\

  LSKNet~\cite{ICCV2023LSKNet}                    & 53.73G            & 30.99M             & 52.39         & 45.15 & 63.59 \\

  \hline
  \multicolumn{5}{l}{\textit{End2end}}  \\
  
  \hline
  DETR~\cite{carion2020end}                 & 24.94G              & 41.56M           & 45.73         & 37.01 & 58.16 \\

  Deformable DETR~\cite{zhu2020deformable}                  & 51.78G & 40.10M     & 52.00  & 46.99 & 63.58 \\
  
  DAB-DETR~\cite{liu2022dab}                  & 28.94G             & 43.70M            & 43.31   & 34.82 & 56.34 \\

  Conditional DETR~\cite{meng2021conditional}                 & 28.09G & 43.45M                         & 44.04         & 35.25 & 56.47 \\
  
  \hline
  \multicolumn{5}{l}{\textit{One-stage}}  \\
  \hline
  FCOS~\cite{tian2019fcos}                       & 51.57G    & 32.13M                     & 52.52        & 47.01 & 66.13 \\
  
  RepPoints~\cite{yang2019reppoints}                 & 48.49G  & 36.82M                       & 51.66         & 46.66 & 63.26 \\
  
  ATSS~\cite{zhang2020bridging}                & 51.57G   & 32.13M                     & 54.95        & 49.89 & 67.94 \\
  
  CenterNet~\cite{zhou2019objects}                   & 51.55G   & 32.12M                      & 53.91         & 48.88 & 66.22\\

  PAA~\cite{kim2020probabilistic}                    & 51.57G  & 32.13M                       & 52.20         & 46.00 & 63.90\\
  
  PVT-T~\cite{wang2021pyramid}                 & 42.19G   & 21.43M                      & 46.10         & 38.01 & 59.53 \\
  
  RetinaNet~\cite{lin2017focal}                 & 52.77G & 36.43M                       & 46.48         & 40.25 & 59.35 \\
  
  TOOD~\cite{feng2021tood}                  & 50.52G & 30.03M                       & 54.65         & 50.20 & 66.72 \\

  DDOD~\cite{ACM2021DDOD}                & 45.58G & 32.21M                       & 54.02         & 49.33 & 64.70 \\
  
  VFNet~\cite{zhang2021varifocalnet}                 & 48.38G & 32.72M                        & 53.01        & 47.37 & 65.39 \\
  
  AutoAssign~\cite{zhu2020autoassign}               & 51.83G  & 36.26M                       & 53.95       & 50.14 & 63.40\\

  YOLOF~\cite{chen2021you}                   & 26.32G & 42.46M                        & 42.83       & 33.73 & 56.19 \\

  YOLOX~\cite{Arixv2021YOLOX}                  & 8.53G  & ~8.94M                       & 34.08       & 28.49 & 43.06 \\
  GFL(w/o Deno)~\cite{li2020generalized}                    & 52.36G  & 32.27M                       & 55.01          & 49.44 & 67.29 \\
  
DenoDet V1~\cite{dai2024denodet}      & 52.69G & 65.78M                            & \underline{55.88}   & \underline{50.63} & \underline{68.47} \\
  \hline

  \rowcolor[rgb]{0.9,0.9,0.9}$\star$ DenoDet V2    & 52.47G & 32.60M                            & \textbf{56.71}   & \textbf{51.45} & \textbf{68.75}\\
  \end{tabular}}
  \vspace{-1em}
  \end{table*}

\begin{table*}[!t]
  \renewcommand\arraystretch{1.2}
  \centering
  \captionsetup{justification=centering}
  \caption{Comparison with SOTA methods on \textbf{AIR-SARShip-1.0} and \textbf{SAR-Aircraft}.}
  \label{tab:sarship}
  \vspace{-2pt}
  \footnotesize
  \setlength{\tabcolsep}{2pt}{
  \begin{tabular}{l|ccaa}
  Method  & \small\textbf{FLOPs $\downarrow$} & \small\textbf{\#P $\downarrow$} & \begin{tabular}[c]{@{}c@{}}\scriptsize{\textbf{AIR-SARShip}}\\\textbf{AP$\uparrow$}\end{tabular} & \begin{tabular}[c]{@{}c@{}} \scriptsize{\textbf{SAR-Aircraft}}\\\textbf{AP$\uparrow$}\end{tabular}  \\
  \Xhline{1pt}
  \multicolumn{5}{l}{\textit{Two-stage}}  \\
  
  \hline
  Faster R-CNN~\cite{ren2015faster}                    & 63.18G  & 41.35M      & 67.48                         &     64.71      \\

  Cascade R-CNN~\cite{cai2019cascade}                  & 90.98G & 69.15M    & 66.69                         &     64.87       \\

  Dynamic R-CNN~\cite{zhang2020dynamic}        & 63.18G & 41.35M        & 67.33                         &            64.59   \\

  Grid R-CNN~\cite{lu2019grid}                    & 180.0G  & 64.47M         & 64.36                       &        64.15    \\
  
  Libra R-CNN~\cite{pang2019libra}                     & 63.99G  & 41.61M      & 67.45                         &       63.46    \\

  ConvNeXt~\cite{CVPR2022ConvNeXt}                       & 63.82G & 45.05M       & 67.52                         &        67.41    \\

  ConvNeXt V2~\cite{CVPR2023ConvNeXtV2}                 & 120.0G & 110.0M        & 69.45                         &     68.04   \\

  LSKNet~\cite{ICCV2023LSKNet}                   & 53.70G & 30.96M      & 71.66                         &     67.58  \\

  \hline
  \multicolumn{5}{l}{\textit{End2end}}  \\
  
  \hline
  DETR~\cite{carion2020end}              & 24.94G    & 41.56M        & 9.09                         &     10.61       \\

  Deformable DETR~\cite{zhu2020deformable}               & 51.77G  & 40.10M      & 55.34             &     62.43      \\
  
  DAB-DETR~\cite{liu2022dab}                     & 28.94G & 43.70M      & 10.72                         &     53.62    \\

  Conditional DETR~\cite{meng2021conditional}             & 28.09G & 43.45M           & 9.09              &    62.25    \\
  \hline
  \multicolumn{5}{l}{\textit{One-stage}}  \\
  \hline
  FCOS~\cite{tian2019fcos}                & 51.50G & 32.11M       & 63.66                         &       62.63  \\
  
  GFL~\cite{li2020generalized}               & 52.30G & 32.26M       & 65.94                         &        66.90   \\
  
  ATSS~\cite{zhang2020bridging}                   & 51.50G  & 32.11M     & 64.21                         &     66.01   \\
  
  CenterNet~\cite{zhou2019objects}               & 51.49G  & 32.11M    & 57.82                         &      64.11    \\

  PAA~\cite{kim2020probabilistic}                & 51.50G    & 32.11M      & 65.65                         &   66.79   \\
  
  PVT-T~\cite{wang2021pyramid}                 & 41.62G    & 21.33M      & 65.59                         &     61.64   \\
  
  RetinaNet~\cite{lin2017focal}                    & 52.20G   & 36.33M        & 65.50                         &   66.47   \\
  
  TOOD~\cite{feng2021tood}                  & 50.46G    & 32.02M      & 63.92                         &   62.66   \\

  DDOD~\cite{ACM2021DDOD}                  & 45.52G  & 32.20M      & 65.29                         &     62.66   \\
  
  VFNet~\cite{zhang2021varifocalnet}            & 48.32G  & 32.71M        & 64.60                         &   66.17    \\
  
  AutoAssign~\cite{zhu2020autoassign}               & 51.77G   & 36.24M       & 65.55                         &    62.36   \\

  YOLOF~\cite{chen2021you}                    & 26.29G    & 42.34M          & 48.32                         &   66.25   \\

  YOLOX~\cite{Arixv2021YOLOX}                 & 8.52G & ~8.94M       & 59.72                         &        63.65    \\
  RepPoints(w/o Deno)~\cite{yang2019reppoints}             & 48.49G  & 36.82M       & 69.88                         &      67.13     \\
      
DenoDet V1~\cite{dai2024denodet}      & 48.52G & 70.33M    & \underline{72.42}           &        \underline{68.60}   \\
  \hline

  \rowcolor[rgb]{0.9,0.9,0.9}$\star$ DenoDet V2    & 48.61G & 37.15M   & \textbf{73.98}      &   \textbf{69.93}    \\
  \end{tabular}}
  \vspace{-1em}
  \end{table*}


\textbf{Results on SARDet-100K: }
As shown in the Tab.~\ref{tab:sardet}, we quantitatively compared our DenoDet V2 with 25 state-of-the-art methods on the highly challenging SARDet-100K dataset, achieving an mAP of \textbf{56.71\%} based on the COCO standard. Notably, compared to its baseline detector GFL, DenoDet V2 improved detection accuracy by \textbf{1.7\%}, and DenoDet V2 achieved the highest mAP of \textbf{51.45\%} and \textbf{68.75\%} for small and medium objects respectively, which are particularly susceptible to noise. Furthermore, the increase in floating-point operations (FLOPs) for DenoDet V2 compared to its base model GFL is almost negligible, the balance of accuracy and efficiency demonstrates the potential of DenoDet V2 for object detection in SAR images.

\textbf{Results on SAR-Aircraft-1.0 and AIR-SARShip-1.0: }
As detailed in Tab. \ref{tab:sarship}, DenoDet V2 sets a new state-of-the-art on both the SAR-Aircraft-1.0 and AIR-SARShip-1.0 benchmarks, outperforming 25 competing methods. It achieves a mAP of 69.93\% on SAR-Aircraft-1.0 and 73.98\% on AIR-SARShip-1.0, surpassing the strong RepPoints baseline by 2.8\% and 4.1\% respectively. These results confirm the model's exceptional capability to handle complex and noisy SAR data, rendering it highly effective for critical applications across military, civilian, and maritime domains.

\subsection{Ablation Study} \label{subsec:ablation}
\textbf{The necessity of phase decomposition: }
To rigorously validate the necessity of orthogonal decomposition of phase angles, a quantitative ablation study was conducted through three controlled experimental configurations as detailed in Tab.~\ref{tab:phase_decompose}. The empirical results reveal that direct angle regression without orthogonal decomposition achieves a baseline performance of 55.6\% mAP on the SARDet-100K benchmark. Implementing orthogonal angular decomposition elevates the mAP to 56.1\%, effectively mitigating periodic boundary discontinuity artifacts inherent in angular parameterization. Subsequent application of trigonometric identity-based angular rectification further improves the mAP to 56.2\%, demonstrating that phase coherence characteristics are better preserved through this geometric regularization process, thereby reducing optimization landscape complexity during model convergence.

\textbf{The Necessity of Phase and Amplitude Refinement: }
Here, we will verify one of the core arguments of this paper: for the frequency domain denoising process of SAR images, the phase spectrum after discrete Fourier transform is more robust compared to the amplitude spectrum and refining both can further improve detection performance. We conducted comparative experiments in Tab.~\ref{tab:transposition}, dividing the experiments into four groups based on whether the phase or amplitude spectrum underwent attention modulation process. From the experimental results, it can be seen that after applying well-designed self-attention to both the phase and amplitude spectrum, the model's detection accuracy improved to some extent. Additionally, on the SARDet-100K dataset, the denoising operation on the phase spectrum improved by another 0.6\% of mAP compared to the amplitude spectrum, proving the viewpoint that the phase information is more robust to noise.


Moreover, signal modulation on both the phase and amplitude spectrum simultaneously can further enhance the detection mAP compared to modulating either one alone. For instance, on the SARDet-100K dataset, the mAP improved by 1.4\% compared to the baseline without denoising and is better than only refining phase or amplitude data. Notably, the detection accuracy for small and medium objects, ${\text{AP}}_{\text{S}}$ and  ${\text{AP}}_{\text{M}}$, increased by 1.5\% and 1.6\% respectively. This indicates that in DenoDet V2, the denoising processes for the phase and amplitude spectrum do not conflict and show promising potential for small and medium objects which are particularly sensitive to noise. This further demonstrates the robustness of DenoDet V2 in denoising.

\setlength{\tabcolsep}{4pt}
\begin{table}[!t]
\centering
\caption{Ablation study on the necessity of phase orthogonal decomposition: impact of phase angular boundary discontinuity issue.}
\label{tab:phase_decompose}
\vspace{-.5\baselineskip}
\scriptsize
\begin{tabular}{c c c c c}
\toprule
\multirow{2}{*}{\begin{tabular}[c]{@{}c@{}}Phase \\ Angle Split \end{tabular}} & \multirow{2}{*}{\begin{tabular}[c]{@{}c@{}}Phase \\ Angle Alignment\end{tabular}} & \multicolumn{3}{c}{SARDet-100K} \\
\cmidrule(lr){3-5}
&  & mAP (COCO) & ${\text{AP}}_{\text{S}}$ &  ${\text{AP}}_{\text{M}}$  \\
\midrule
\xmark & \xmark & 55.6 & 49.5 & 68.5 \\
$\checkmark$ & \xmark & 56.1 & 50.4 & 68.6 \\
\rowcolor[rgb]{0.9,0.9,0.9} $\checkmark$ & $\checkmark$  & \textbf{56.2} & \textbf{51.4} & \textbf{68.6}\\
\bottomrule
\end{tabular}
\end{table}

\setlength{\tabcolsep}{4pt}
\begin{table}[!t]
\centering
\caption{Ablation study on the necessity of phase and amplitude refine: phase spectrum vs amplitude spectrum.}
\label{tab:transposition}
\vspace{-.5\baselineskip}
\scriptsize
\begin{tabular}{c c c c c}
\toprule
\multirow{2}{*}{\begin{tabular}[c]{@{}c@{}}DFT \\ Amplitude Refine\end{tabular}} & \multirow{2}{*}{\begin{tabular}[c]{@{}c@{}}DFT \\ Phase Refine\end{tabular}} & \multicolumn{3}{c}{SARDet-100K} \\
\cmidrule(lr){3-5}
&  & mAP (COCO) & ${\text{AP}}_{\text{S}}$ &  ${\text{AP}}_{\text{M}}$  \\
\midrule

\xmark & \xmark & 55.0 & 49.4 & 67.3 \\


$\checkmark$ & \xmark & 55.6 & 50.3 & 68.2 \\

\xmark & $\checkmark$ & 56.2 & 51.4 & 68.6 \\

\rowcolor[rgb]{0.9,0.9,0.9} $\checkmark$ & $\checkmark$  & \textbf{56.4} & \textbf{50.9} & \textbf{68.9}\\
\bottomrule
\end{tabular}
\end{table}

\setlength{\tabcolsep}{6pt}
\begin{table}[!t]
\centering
\caption{Ablation study on the necessity of band partition: impact of band partition stride.}
\label{tab:Partition_Stride}
\vspace{-.5\baselineskip}
\small
\begin{tabular}{c c c c c }
\toprule 
\multicolumn{2}{c}{\multirow{2}{*}{\begin{tabular}[c]{@{}c@{}}Partition \\ Stride\end{tabular}}} & \multicolumn{3}{c}{SARDet-100K} \\
\cmidrule(lr){3-5}
&  &mAP (COCO) & ${\text{AP}}_{\text{S}}$ &  ${\text{AP}}_{\text{M}}$\\
\midrule
 & 1 & 52.5 & 45.6 & 63.8\\
  & 2 & 55.7 & 49.5 & 68.7 \\
  & 4 & 56.2 & 51.3 & 68.5 \\
  \rowcolor[rgb]{0.9,0.9,0.9} & 8 & \textbf{56.7} & \textbf{51.5} & \textbf{68.8} \\
  & 16 & 56.1 & 51.0 & 68.2 \\
\bottomrule
\end{tabular}
\end{table}

\setlength{\tabcolsep}{6pt}
\begin{table}[!t]
\centering
\caption{Ablation study on the necessity of token exchange: impact of token exchange model design.}
\label{tab:Token_Exchange}
\vspace{-.5\baselineskip}
\small
\begin{tabular}{l l c c c}
\toprule
\multicolumn{2}{c}{\multirow{2}{*}{Design}} & \multicolumn{3}{c}{SARDet-100K} \\
\cmidrule(lr){3-5}
&  & mAP (COCO) & ${\text{AP}}_{\text{S}}$ &  ${\text{AP}}_{\text{M}}$ \\
\midrule
 & baseline & 55.0 & 49.4 & 67.3 \\
 & No-Exchange & 56.4 & 50.9 & \textbf{68.9} \\
  \rowcolor[rgb]{0.9,0.9,0.9} & Token-Exchange & \textbf{56.7} & \textbf{51.5} & 68.8 \\
\bottomrule
\end{tabular}
\vspace*{-1.\baselineskip}
\end{table}


\textbf{The Necessity of Band Partition: }
We verify the necessity of band partition and discuss the impact of different band partition strides. Based on the 2D frequency characteristics of signals from the DFT, the proposed BPSA module groups and partitions frequency domain data to ensure that the frequency variation within each group is uniform, decoupling the signal modulation process, reducing the model's convergence difficulty and parameter count. We conducted comparative experiments in Tab.~\ref{tab:Partition_Stride}, dividing the experiments into five groups based on the stride size of the band partition.

From the experimental results, it can be seen that when the stride is 1, which means band partition is discarded, the model's performance is the worst, and the parameter count significantly increases. It proves the necessity of band separation operation. As the partition stride increases, the model's detection accuracy also improves, with the best performance observed at a stride of 8, while results of 16 stride in a slight decrease. In this case, in the proposed DenoDet V2, a stride of 8 is adopted for the final experimental results.


\textbf{The Necessity of Token Exchange: }
Finally, we conducted ablation experiments to demonstrate the effectiveness of the token exchange strategy. As shown in the Tab.~\ref{tab:Token_Exchange}, we found that after applying the token exchange operation, the mAP on the SARDet-100K dataset increased from 56.4\% to 56.7\%. This indicates that the token exchange operation facilitates better interaction between amplitude and phase information, achieving improved modulation effects. Furthermore, compared to the baseline model, our implementation incorporating token exchange between amplitude and phase spectrum achieves a 1.7\% mAP improvement, with a 2.1\% AP enhancement for small objects. This performance gain demonstrates DenoDetV2's effectiveness in suppressing specific speckle noise in SAR images through cross-spectral feature recalibration while enhancing object saliency, ultimately improving detection fidelity in cluttered environments.

\subsection{Visual Analysis} \label{sec:analysis}
\begin{figure}[!t]
  \centering
  \includegraphics[width=0.5\textwidth]{"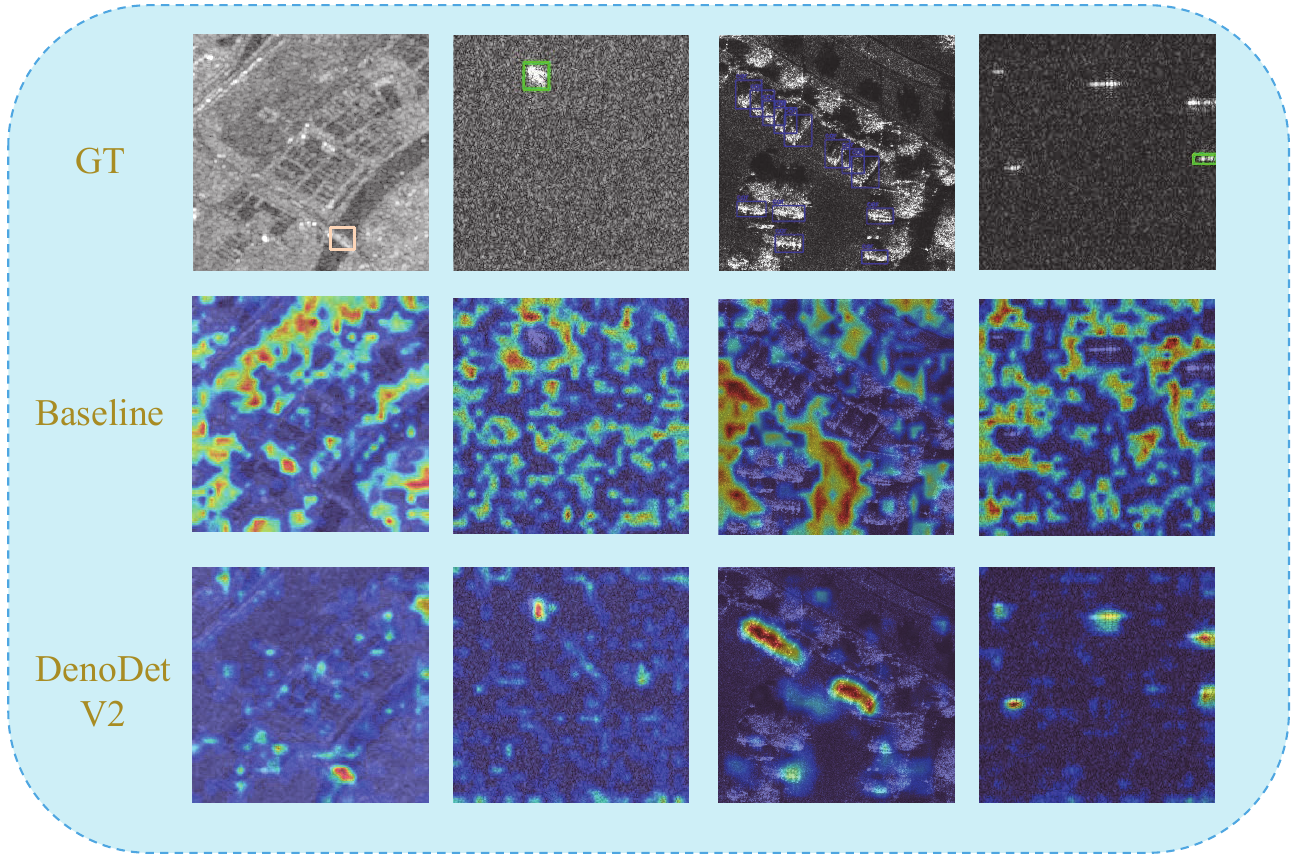"}
  \vspace*{-1.\baselineskip}
  \caption{Feature heatmap visualization on SARDet-100K. Compared to the baseline, DenoDet V2 focuses more attention on the object areas while being less affected by the background noise.}
  \label{fig:heatmap}
  \vspace*{-1.\baselineskip}
\end{figure}

In this part, we provide visual analysis to support our core hypothesis, utilizing Eigen-CAM-based visualizations to demonstrate the effectiveness of our DFTDeno module. From Fig. \ref{fig:heatmap}, it is evident that our proposed DenoDet V2 model, compared to its original baseline model, exhibits more vibrant colors in the object area. Additionally, the highlighted regions are concentrated near the objects, while the attention dispersed in noise-affected areas is significantly suppressed.



\begin{figure}[!t]
  \centering
  \includegraphics[width=0.5\textwidth]{"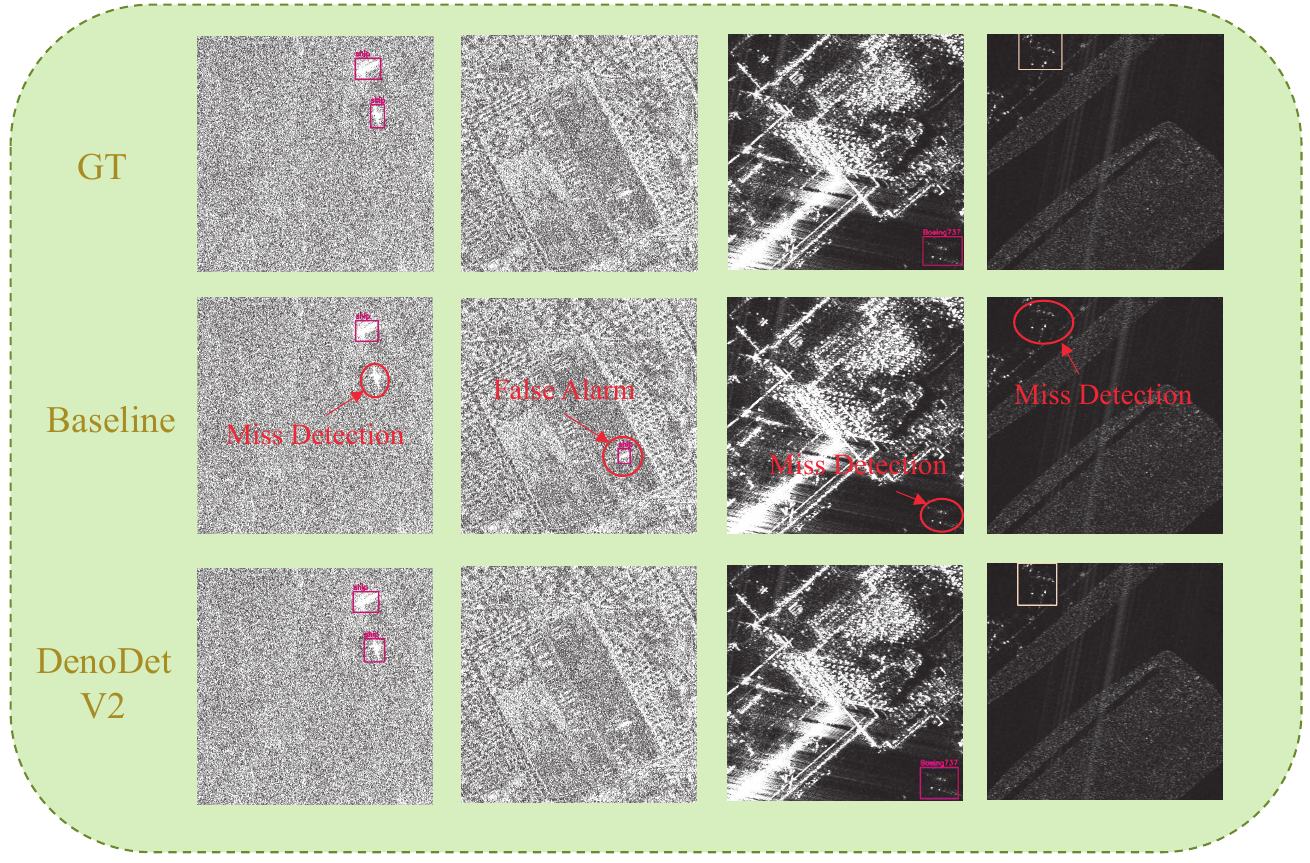"}
  \vspace*{-1.\baselineskip}
  \caption{Detection result comparison on SAR-AIRCraft-1.0 and AIR-SARShip-1.0 dataset. DenoDet V2 achieved more accurate detection results in noisy regions and identified more ship objects.}
  \label{fig:detection_results}
  \vspace*{-1.\baselineskip}
\end{figure}

Additionally, we visualized the detection results on SAR-AIRCraft-1.0 and AIR-SARShip-1.0, as shown in Fig. \ref{fig:detection_results}. Even under severe noise interference, DenoDet V2 can still accurately detect ship targets without generating false alarms caused by noise. This further demonstrates the robustness of our proposed method against noise.





\section{Conclusion} \label{sec:conclusion}

In this paper, we presented DenoDet V2, an approach for robust SAR image object detection that leverages the complementary nature of amplitude and phase information through a mutual guidance mechanism. Our method introduces the concept of Attention as Phase-Amplitude Denoising, which exploits the inherent stability of phase information to guide the feature denoising of amplitude in SAR imagery. The Phase-Amplitude Token Exchange (PATE) module, a key component of DenoDet V2, enables a reciprocal enhancement between phase and amplitude spectra, resulting in a synergistic feature denoising effect that iteratively boosts the accuracy of object detection.
Through extensive experiments on various SAR datasets, we demonstrated the state-of-the-art performance of DenoDet V2, showcasing its potential for robust object detection in noisy SAR environments. 

\clearpage


{
    \small
    \bibliographystyle{ieeenat_fullname}
    \bibliography{main}
}

\clearpage
\setcounter{page}{1}
\renewcommand{\thetable}{S\arabic{table}}
\renewcommand{\thefigure}{S\arabic{figure}}

\onecolumn
\begin{center}
\Large
\vspace{0.5em}Supplementary Material \\
\vspace{1.0em}
\normalsize 
\end{center}
    
\section*{\textbf{Results on SARDet-100K}}
\begin{figure*}[!h]
  \centering
  \includegraphics[width=1\textwidth]{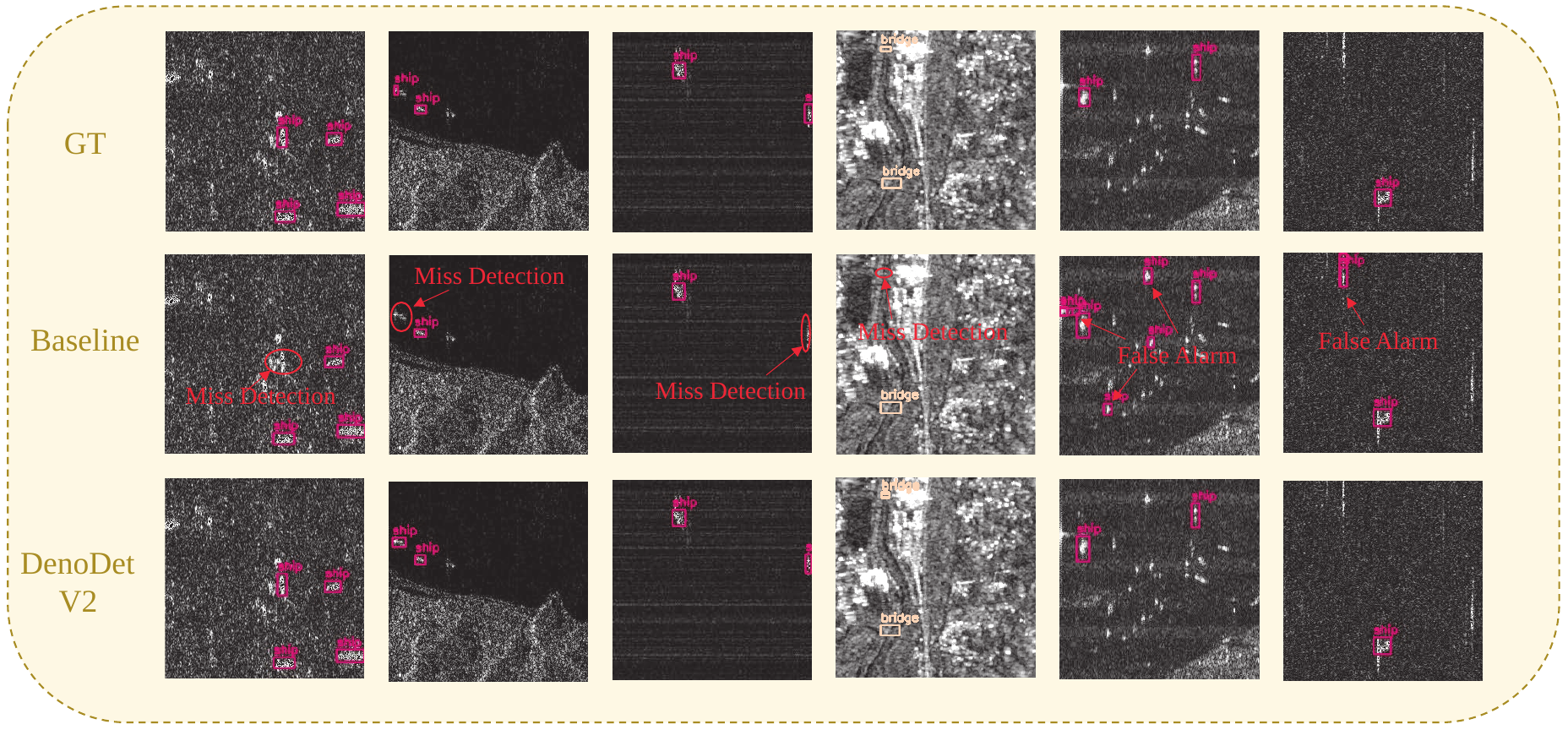}
  \caption{Comparative visualization results on SARDet-100K dataset. We highlight the missed detections or false alarms of the baseline model with circles. DenoDet V2 achieves excellent detection performance in both noisy images and complex scenes with multiple objects.}
  \label{fig:sardet_sup}
\end{figure*}

As shown in Tab. \ref{tab:sardet_full_sup}, our DenoDetV2 achieves state-of-the-art performance on the SARDet-100K dataset with 56.71\% mAP under the COCO evaluation protocol, outperforming all 25 compared methods including both classical two-stage detectors and modern transformer-based approaches. Notably, it surpasses its baseline GFL by 1.7\% mAP, while maintaining nearly identical computational complexity, demonstrating exceptional noise robustness through 51.45\% AP for small objects and 68.75\% AP for medium objects. In addition, we visualized the detection results of DenoDet V2 on the SARDet-100K dataset, as shown in the Fig. \ref{fig:sardet_sup}. In areas with noise interference, DenoDet V2 produced fewer false alarms and detected more targets. Even in more complex and crowded regions, DenoDet V2 demonstrated superior detection performance.

\begin{table*}[h]
  \renewcommand\arraystretch{1.2}
  \centering
  \vspace{-2pt}
  \captionsetup{justification=centering}
  \caption{Comparison with SOTA methods on the \textbf{SARDet-100K} dataset.}
  \label{tab:sardet_full_sup}
  \scriptsize
  \setlength{\tabcolsep}{1pt}{
  \resizebox{\columnwidth}{!}{
  \begin{tabular}{l|c|cc|accccccccccc}
  Method & \small Pre. & \small\textbf{FLOPs $\downarrow$} & \small\textbf{\#P $\downarrow$} & \small\textbf{mAP} & \small\textbf{AP@50}  & \small\textbf{AP@75}  & $\small\textbf{AP}_{\small\textbf{S}}$  & $\small\textbf{AP}_{\small\textbf{M}}$  & $\small\textbf{AP}_{\small\textbf{L}}$ & \small\textbf{Ship} & \small\textbf{Aircraft} & \small\textbf{Car} & \small\textbf{Tank} & \small\textbf{Bridge} & \small\textbf{Harbor}\\
  \Xhline{1pt}
    \multicolumn{5}{l}{\textit{Two-stage}}  \\
  
  \hline
  Faster R-CNN                & IN       & 63.2G  & 41.37M                       & 39.22         & 70.04 & 39.87 & 32.55 & 47.23 & 42.02 & 50.45 & 50.36 & 57.82 & 24.90 & 18.69 & 33.11                 \\

  Cascade R-CNN                & IN       & 90.99G     & 69.17M                    & 53.55         & 87.33 & 56.81 & 49.09 & 62.89 & 48.68 & \textbf{66.99} & 56.43 & 63.25 & 44.35 & 36.89 & 53.81                \\

  Dynamic R-CNN                & IN       & 63.2G      & 41.37M                   & 49.75         & 80.96 & 53.91 & 43.12 & 59.72 & 54.77 & 61.32 & 53.86 & 60.00 & 33.68 & 34.40 & 55.25                 \\

  Grid R-CNN                & IN       & 0.18T            & 64.47M             & 50.05         & 80.58 & 53.49 & 42.43 & 62.01 & 52.70 & 60.43 & 55.61 & 61.94 & 36.03 & 31.16 & 55.13                \\
  
  Libra R-CNN                & IN       & 64.02G       & 41.64M                  & 52.09         & 83.54 & 55.81 & 45.85 & 63.52 & 55.40 & 61.32 & 54.03 & 61.56 & 38.12 & 35.97 & 61.50                 \\

  ConvNeXt                & IN       & 63.84G          & 45.07M              & 53.15         & 85.52 & 57.28 & 45.67 & 64.55 & 58.61 & 60.55 & 57.35 & 62.13 & 38.12 & 36.81 & 63.95                 \\

  ConvNeXtV2                & IN       & 0.12T         & 0.11G                & 53.91         & 86.01 & 58.90 & 47.63 & 64.67 & 59.57 & 61.48 & 55.83 & 63.23 & 39.65 & \textbf{39.16} & 64.09                 \\

  LSKNet                & IN       & 53.73G            & 30.99M             & 52.39         & 85.07 & 56.96 & 45.15 & 63.59 & 59.16 & 59.33 & 56.76 & 62.74 & 36.09 & 35.01 & 64.38                 \\

  \hline
  \multicolumn{5}{l}{\textit{End2end}}  \\
  
  \hline
  DETR                & IN       & 24.94G              & 41.56M           & 45.73         & 78.57 & 46.87 & 37.01 & 58.16 & 55.58 & 54.94 & 51.17 & 50.11 & 26.06 & 32.80 & 59.31                 \\

  Deformable DETR                & IN       & 51.78G & 40.10M                        & 52.00         & \underline{88.77} & 54.03 & 46.99 & 63.58 & 58.55 & 60.94 & 54.16 & 61.22 & 39.14 & 36.09 & 60.46               \\
  
  DAB-DETR                & IN       & 28.94G             & 43.70M            & 43.31         & 78.14 & 43.10 & 34.82 & 56.34 & 52.62 & 53.16 & 50.32 & 49.47 & 24.06 & 28.47 & 55.07                 \\

  Conditional DETR                & IN       & 28.09G & 43.45M                         & 44.04         & 77.88 & 44.40 & 35.25 & 56.47 & 52.86 & 52.77 & 49.58 & 51.00 & 22.73 & 29.98 & 58.16                 \\
  
  \hline
  \multicolumn{15}{l}{\textit{One-stage}}  \\
  \hline
  FCOS                  & IN       & 51.57G    & 32.13M                     & 52.52        & 85.82 & 54.93 & 47.01 & 66.13 & 57.82 & 59.79 & 55.44 & 60.75 & 41.78 & 34.17 & 63.44               \\
  
  GFL                & IN       & 52.36G  & 32.27M                       & 55.01         & 85.16 & 58.87 & 49.44 & 67.29 & 60.45 & 63.92 & \underline{57.63} & 62.29 & 44.80 & 36.41 & 65.04                 \\
  
  RepPoints                & IN       & 48.49G  & 36.82M                       & 51.66         & 86.43 & 53.99 & 46.66 & 63.26 & 53.78 & 60.85 & 55.50 & 61.13 & 40.69 & 35.12 & 56.71                 \\
  
  ATSS                & IN       & 51.57G   & 32.13M                     & 54.95        & 87.60 & 58.25 & 49.89 & 67.94 & 58.97 & 61.53 & 55.94 & 61.77 & \underline{46.20} & 37.22 & \underline{67.48}               \\
  
  CenterNet                & IN       & 51.55G   & 32.12M                      & 53.91         & 86.17 & 57.31 & 48.88 & 66.22 & 57.74 & 61.24 & 56.35 & 61.74 & 45.31 & 35.91 & 63.29                 \\

  PAA                & IN       & 51.57G  & 32.13M                       & 52.20         & 85.71 & 54.80 & 46.00 & 63.90 & 57.61 & 60.16 & 56.17 & 60.09 & 41.07 & 35.96 & 60.12               \\
  
  PVT-T                & IN       & 42.19G   & 21.43M                      & 46.10         & 77.55 & 49.00 & 38.01 & 59.53 & 53.35 & 53.30 & 52.91 & 59.03 & 30.20 & 22.51 & 59.11                 \\
  
  RetinaNet                & IN       & 52.77G & 36.43M                       & 46.48         & 77.74 & 48.94 & 40.25 & 59.35 & 50.26 & 55.36 & 54.00 & 60.88 & 32.72 & 24.81 & 51.12                 \\
  
  TOOD                & IN       & 50.52G & 30.03M                       & 54.65         & 86.88 & 58.41 & 50.20 & 66.72 & 58.60 & 62.28 & 55.61 & 62.53 & 45.96 & 36.64 & 65.24                 \\

  DDOD                & IN       & 45.58G & 32.21M                       & 54.02         & 86.64 & 57.23 & 49.33 & 64.70 & 58.02 & 62.39 & 56.08 & 62.48 & 43.98 & 36.34 & 62.87                 \\
  
  VFNet                & IN       & 48.38G & 32.72M                        & 53.01         & 84.32 & 56.32 & 47.37 & 65.39 & 57.99 & 62.14 & 55.84 & 61.97 & 42.08 & 34.11 & 62.28                 \\
  
  AutoAssign                & IN       & 51.83G  & 36.26M                       & 53.95        & \textbf{89.58} & 55.96 & 50.14 & 63.40 & 54.73 & 62.03 & 55.70 & 61.69 & \textbf{48.55} & \underline{38.25} & 57.45                 \\

  YOLOF                & IN       & 26.32G & 42.46M                        & 42.83        & 74.95 & 43.18 & 33.73 & 56.19 & 53.57 & 52.62 & 52.64 & 52.71 & 22.86 & 23.74 & 52.42                 \\

  YOLOX                & IN       & 8.53G  & 8.94M                       & 34.08        & 66.77 & 31.31 & 28.49 & 43.06 & 28.95 & 46.08 & 46.83 & 53.43 & 26.26 & 13.14 & 18.95                 \\
  
  \hline
  \rowcolor[rgb]{0.9,0.9,0.9}$\star$ DenoDet    & IN     & 52.69G & 65.78M                            & \underline{55.88}   & 85.81 & \underline{60.16} & \underline{50.63} & \underline{68.47} & \underline{60.96} & \underline{64.91} & 57.36 & \textbf{63.66} & 45.79 & 36.39 & 67.17  \\
    
  \rowcolor[rgb]{0.9,0.9,0.9}$\star$ DenoDetV2    & IN     & 52.47G & 32.60M                            & \textbf{56.71}   & 85.86 & \textbf{61.55} & \textbf{51.45} & \textbf{68.75} & \textbf{61.34} & 64.85 & \textbf{58.08} & \underline{63.26} & 46.18 & 37.66 & \textbf{70.22}   \\
  \end{tabular}}}
  \end{table*}

\section*{\textbf{Results on SAR-Aircraft-1.0}}
\begin{figure*}[!h]
  \centering
  \includegraphics[width=1\textwidth]{"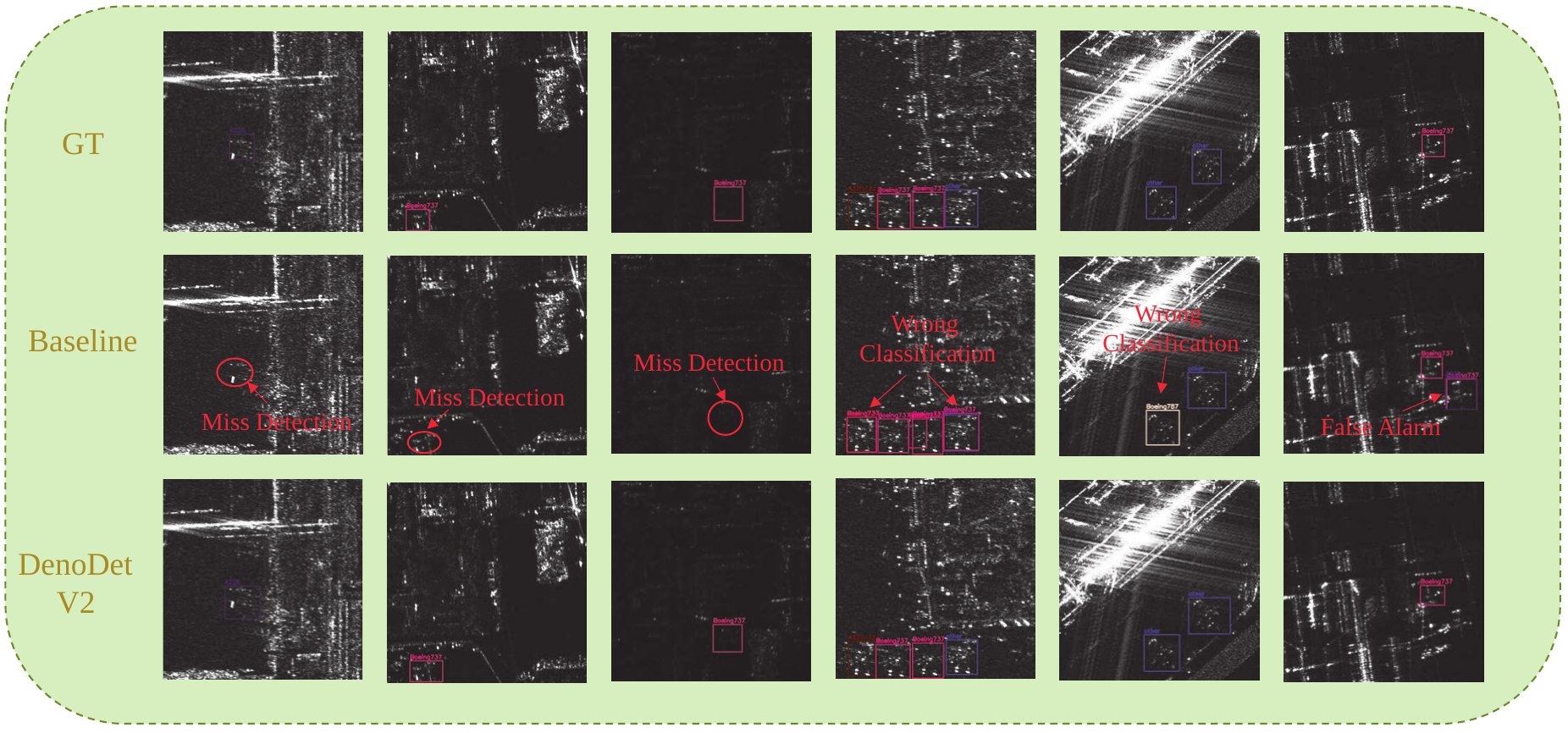"}
  \caption{Comparative visualization results on SAR-AIRcraft-1.0 dataset. The missed detections, false alarms, and misclassified objects of the baseline model are marked with red circles. It can be observed that aircraft objects such as the Boeing 787 and A220 are correctly detected and classified, indicating that DenoDet V2 demonstrates excellent performance when generalized to a broader range of detection scenarios.}
  \label{fig:saraircraft_sup}
\end{figure*}

The comprehensive comparison in Tab. ~\ref{tab:aircraft_sup} highlights DenoDetV2's exceptional performance on the SAR-Aircraft-1.0 dataset, achieving state-of-the-art results with 69.93\% mAP for AP'07 and 70.73\% mAP for AP'12. It outperforms all 25 benchmarked methods across various architectural paradigms. DenoDetV2 makes significant strides in detecting challenging aircraft variants, achieving 49.06\% AP'07 for B737 targets, a 2.9\% improvement over GFL, and 68.02\% AP'07 for B787. The model maintains parameter efficiency with only 37.18 million parameters, 47\% fewer than its predecessor DenoDetV1, which had 70.33 million parameters, while also improving mAP by 1.33\%, demonstrating optimized feature encoding. After visualizing the detection results on the SAR-Aircraft-1.0 dataset in the Fig. \ref{fig:saraircraft_sup}, we find that DenoDet V2 exhibits strong detection performance in noisy images. At the same time, it still achieves excellent detection results in cleaner SAR images, demonstrating the generalization ability of DenoDet V2 in detection tasks.

  \begin{table*}[h]
    \renewcommand\arraystretch{1.2}
    \captionsetup{justification=centering}
    \caption{Comparison with SOTA methods on the \textbf{SAR-Aircraft-1.0} dataset.}
    \label{tab:aircraft_sup}
    \scriptsize
    \centering
    \vspace{-2pt}
    \setlength{\tabcolsep}{1pt}
    \begin{tabular}{l|c|cc|accccccc|accccccc}
      \multirow{2}{*}{Method} & \multirow{2}{*}{Pre.}     & \multirow{2}{*}{FLOPs $\downarrow$} & \multirow{2}{*}{\#P $\downarrow$} & \multicolumn{8}{c|}{Average Precision (AP'07)}  & \multicolumn{8}{c}{Average Precision (AP'12)} \\
        &  & &  & mAP &  A220  & A320    &  A330   & ARJ21 & B737 & B787 & Other & mAP & A220  & A320    &  A330   & ARJ21 & B737 & B787 & Other \\
    \Xhline{1pt}
        \multicolumn{19}{l}{\textit{Two-stage}}  \\
    \hline
  
    Faster R-CNN & IN    & 63.21G & 41.38M & 64.71 & 56.96 & 88.09 & 96.99 & 52.78 & 35.74 & 59.08 & 63.32 & 65.11 & 58.63 & 90.53 & 97.47 & 56.82 & 32.31 & 57.07 & 62.94 \\
  
    Cascade R-CNN & IN    & 91.00G & 69.17M & 64.87 & 51.87 & \underline{96.32} & \underline{97.51} & \underline{62.65} & 30.27 & 58.39 & 57.09 & 65.27 & 52.76 & 96.58 & 97.73 & 65.66 & 27.21 & 56.10 & 60.88 \\
  
    Dynamic R-CNN & IN    & 63.21G & 41.38M & 64.59 & 51.03 & 92.49 & \textbf{97.56} & 59.63 & 32.37 & 56.21 & 62.82 & 65.38 & 53.71 & 93.23 & \textbf{98.02} & 59.03 & 32.07 & 58.86 & 62.72 \\
  
    Grid R-CNN & IN    & 0.18T & 64.47M & 64.15 & 54.05 & \textbf{98.08} & 97.16 & 60.47 & 28.94 & 47.73 & 62.61 & 64.04 & 54.48 & \underline{98.42} & 97.36 & 59.54 & 26.80 & 49.14 & 62.51 \\
  
    Libra R-CNN & IN    & 64.02G & 41.64M & 63.46 & 55.47 & 94.35 & 88.72 & 53.63 & 37.03 & 52.90 & 62.12 & 65.47 & 56.27 & 96.79 & 93.63 & 57.12 & 35.03 & 54.89 & 64.55 \\

    ConvNeXt & IN    & 63.85G & 45.08M & 67.41 & 63.44 & 92.25 & 97.53 & 69.35 & 38.58 & 49.06 & 61.64 & 66.98 & 63.51 & 93.71 & 97.66 & \textbf{67.61} & 36.17 & 49.06 & 61.11 \\
    
    ConvNeXt V2 & IN    & 0.12T & 0.11G & 68.04 & 61.13 & 93.77 & 97.16 & 59.73 & 36.28 & \underline{62.38} & 65.82 & 68.72 & 62.71 & 94.88 & 97.30 & 59.93 & 34.24 & \underline{63.19} & \underline{68.76} \\
    
    LSKNet & IN    & 53.73G & 30.99M & 67.58 & 59.15 & 99.06 & 96.83 & 58.85 & 37.34 & 56.84 & 64.98 & 68.26 & 59.82 & \textbf{99.21} & 97.21 & 60.80 & 35.56 & 57.47 & 67.79 \\

    \hline
    \multicolumn{19}{l}{\textit{End2end}}  \\
    \hline
    DETR & IN    & 24.94G & 41.56M & 10.61 & 19.55 & 0.78 & 0.47 & 4.15 & 9.11 & 20.95 & 19.25 & 8.27 & 15.80 & 0.56 & 0.45 & 2.45 & 6.14 & 18.97 & 13.55 \\
    Deformable DETR & IN    & 51.78G & 40.10M & 62.43 & \textbf{66.76} & 69.52 & 89.10 & 59.17 & 40.98 & 48.62 & 62.86 & 63.61 & \textbf{68.35} & 71.74 & 94.05 & 60.68 & 38.82 & 47.83 & 63.81 \\
    
    DAB-DETR & IN    & 28.94G & 43.70M & 53.62 & 59.82 & 88.39 & 18.84 & 62.26 & 34.52 & 54.34 & 57.15 & 54.16 & 60.80 & 91.10 & 18.97 & 63.61 & 33.20 & 54.04 & 57.42 \\
  
    Conditional DETR & IN    & 28.09G & 43.45M & 62.25 & 53.74 & 89.38 & 78.83 & 61.75 & 37.44 & 61.45 & 53.19 & 63.02 & 54.27 & 93.30 & 80.47 & 62.07 & 35.92 & 61.99 & 53.14 \\
    \hline
    \multicolumn{19}{l}{\textit{One-stage}}  \\
    \hline
  
    FCOS & IN    & 51.58G & 32.13M & 62.63 & 62.27 & 93.19 & 63.65 & \textbf{63.10} & 40.06 & 52.37 & 63.79 & 63.76 & 64.63 & 95.16 & 64.61 & 63.02 & 39.36 & 53.05 & 66.50 \\
  
    GFL & IN    & 52.37G & 32.27M & 66.90 & 59.90 & 87.34 & 92.88 & 62.53 & 46.16 & 61.82 & 57.64 & 68.44 & 61.19 & 90.61 & 94.75 & \underline{66.10} & \underline{45.98} & 62.11 & 58.38 \\
  
    RepPoints & IN    & 48.50G & 36.82M & 67.13 & 64.27 & 89.75 & 86.28 & 61.28 & 41.50 & 59.73 & \underline{67.10} & 68.09 & 65.72 & 93.03 & 86.91 & 62.88 & 40.01 & 59.80 & 68.29 \\
  
    ATSS & IN    & 51.58G & 32.13M & 66.01 & 58.57 & 88.60 & \underline{96.57} & 60.14 & 37.79 & 59.18 & 61.25 & 66.71 & 59.84 & 92.03 & 96.74 & 61.78 & 36.12 & 59.10 & 61.36 \\
  
    CenterNet & IN    & 51.57G & 32.13M & 64.11 & 58.03 & 93.03 & \textbf{97.02} & 61.90 & 36.41 & 41.74 & 60.61 & 64.90 & 58.15 & 96.58 & 97.36 & 62.13 & 34.99 & 42.28 & 62.81 \\
  
    PAA & IN    & 51.58G & 32.13M & 66.79 & \underline{66.26} & 89.56 & 96.41 & 62.20 & 35.90 & 54.10 & 63.12 & 67.56 &        \underline{67.56} & 93.51 & 97.17 & 62.91 & 34.69 & 53.04 & 64.07 \\
  
    PVT-T & IN    & 42.30G & 21.45M & 61.64 & 52.70 & 82.01 & 85.94 & 61.07 & 32.63 & 54.94 & 62.15 & 62.43 &        53.32 & 84.46 & 87.85 & 62.12 & 30.57 & 55.56 & 63.15 \\
  
    RetinaNet & IN    & 52.88G & 36.46M & 66.47 & 65.77 & 95.10 & 82.31 & 54.79 & 41.22 & 60.59 & 65.52 & 67.26 & 66.66 & 97.09 & 84.08 & 54.37 & 39.33 & 61.16 & 68.10 \\
  
    TOOD & IN    & 50.53G & 32.03M & 62.66 & 53.94 & 82.20 & 91.86 & 60.51 & 35.36 & 53.81 & 60.95 & 62.98 & 53.81 & 83.91 & 93.99 & 60.51 & 33.91 & 53.50 & 61.21 \\

    DDOD & IN    & 45.60G & 32.21M & 62.66 & 53.02 & 87.34 & 96.12 & 59.95 & 35.31 & 44.54 & 62.39 & 63.10 & 52.44 & 91.45 & 96.31 & 61.40 & 34.83 & 42.40 & 62.87 \\
  
    VFNet & IN    & 48.39G & 32.72M & 66.17 & 60.94 & 90.91 & 95.25 & 61.29 & 39.41 & 56.46 & 58.93 & 66.90 &  62.00 & 95.57 & 96.14 & 60.76 & 36.63 & 56.76 & 60.43 \\
    
    AutoAssign & IN    & 51.84G & 36.26M & 62.36 & 59.40 & 82.98 & 92.62 & 57.49 & 32.62 & 51.21 & 60.18 & 63.11 & 60.61 & 86.39 & 94.02 & 57.87 & 30.86 & 51.31 & 60.74 \\

    YOLOF & IN    & 26.33G & 42.48M & 66.25 & 60.24 & 90.75 & 88.81 & 61.00 & 36.30 & 59.64 & 66.99 & 67.71 & 60.97 & 94.85 & 91.31 & 63.37 & 34.35 & 60.42 & 68.71 \\

    YOLOX & IN    & 8.53G & 8.94M & 63.65 & 63.93 & 89.61 & 97.54 & 52.01 & 24.19 & 56.63 & 61.67 & 65.05 & 64.91 & 91.27 & \underline{97.88} & 55.51 & 22.88 & 60.65 & 62.26 \\

    \hline
    \rowcolor[rgb]{0.9,0.9,0.9}$\star$ DenoDet  & IN    & 48.53G & 70.33M & \underline{68.60}   &  64.82 & 90.54 & 93.82 & 59.58 & \underline{46.32} & 56.04 & \textbf{69.05} & \underline{69.56} & 66.53 & 93.43 & 94.84 & 60.80 & 44.85 & 55.74 & \textbf{70.71} \\
        
    \rowcolor[rgb]{0.9,0.9,0.9}$\star$ DenoDetV2  & IN    & 48.61G & 37.15M & \textbf{69.93}   &  60.52 & 91.80 & 97.21 & 56.13 & \textbf{49.06} & \textbf{68.02} & 66.76 & \textbf{70.73} & 61.46 & 93.42 & 97.56 & 57.06 & \textbf{47.84} & \textbf{69.99} & 67.76 \\
    \end{tabular}
    \end{table*}

\begin{table*}[!h]
  \captionsetup{justification=centering}
  \caption{Comparison with SOTA methods on the \textbf{AIR-SARShip-1.0} dataset.}
  \label{tab:sarship_full_sup}
  \renewcommand\arraystretch{1.2}
  \centering
  \vspace{-2pt}
  \scriptsize
  \setlength{\tabcolsep}{1mm}{
  \resizebox{0.6 \columnwidth}{!}{
  \begin{tabular}{lcaacc}
  Method & \small Pre. & \small\textbf{FLOPs $\downarrow$} & \small\textbf{\#P $\downarrow$} & \small\textbf{AP(07)$\uparrow$} & \small\textbf{AP(12)$\uparrow$}\\
  \Xhline{1pt}
  \multicolumn{5}{l}{\textit{One-stage}}  \\
  \hline
  FCOS                  & IN       & 51.50G & 32.11M & 63.66                         & 64.25                        \\
  
  GFL                & IN       & 52.30G & 32.26M & 65.94                         & 66.05                         \\
  
  RepPoints                & IN  & 48.49G  & 36.82M      & 69.88                         & 70.27                       \\
  
  ATSS                & IN       & 51.50G  & 32.11M & 64.21                         & 64.87                       \\
  
  CenterNet                & IN       & 51.49G  & 32.11M & 57.82                         & 57.55                        \\

  PAA                & IN       & 51.50G    & 32.11M & 65.65                         & 66.96                      \\
  
  PVT-T                & IN       & 41.62G    & 21.33M & 65.59                         & 65.94                      \\
  
  RetinaNet                & IN      & 52.20G   & 36.33M & 65.50                         & 65.60                       \\
  
  TOOD                & IN       & 50.46G    & 32.02M & 63.92                         & 64.37                      \\

  DDOD                & IN       & 45.52G  & 32.20M & 65.29                         & 65.45                        \\
  
  VFNet                & IN       & 48.32G  & 32.71M & 64.60                         & 66.62                        \\
  
  AutoAssign                & IN        & 51.77G   & 36.24M & 65.55                         & 66.05                      \\

  YOLOF                & IN      & 26.29G    & 42.34M & 48.32                         & 49.43                      \\

  YOLOX                & IN       & 8.52G & 8.94M & 59.72                         & 62.35                          \\
  
  \hline
  \multicolumn{5}{l}{\textit{Two-stage}}  \\
  
  \hline
  Faster R-CNN                & IN & 63.18G  & 41.35M       & 67.48                         & 66.14                        \\

  Cascade R-CNN                & IN & 90.98G & 69.15M      & 66.69                         & 65.67                         \\

  Dynamic R-CNN                & IN & 63.18G & 41.35M      & 67.33                         & 66.23                         \\

  Grid R-CNN                & IN      & 0.18T  & 64.47M & 64.36                         & 66.12                        \\
  
  Libra R-CNN                & IN       & 63.99G  & 41.61M & 67.45                         & 70.30                        \\

  ConvNeXt                & IN       & 63.82G & 45.05M & 67.52                         & 67.82                         \\

  ConvNeXt V2                & IN       & 0.12T & 0.11G & 69.45                         & 70.52                         \\

  LSKNet                & IN       & 53.70G & 30.96M & 71.66                         & 72.02                        \\

  \hline
  \multicolumn{5}{l}{\textit{End2end}}  \\
  
  \hline
  DETR               & IN       & 24.94G    & 41.56M & 9.09                         & 0.83                      \\

  Deformable DETR                & IN & 51.77G  & 40.10M      & 55.34                         & 54.68                        \\
  
  DAB-DETR                & IN     & 28.94G & 43.70M  & 10.72                         & 5.42                         \\

  Conditional DETR                & IN    & 28.09G & 43.45M   & 9.09                         & 0.84                         \\
  
  \hline
  \rowcolor[rgb]{0.9,0.9,0.9}$\star$ DenoDet    & IN        & 48.52G & 70.33M  & \underline{72.42}                              &              \underline{73.36} \\
  \rowcolor[rgb]{0.9,0.9,0.9}$\star$ DenoDetV2    & IN     & 48.61G & 37.15M  & \textbf{73.98}                              &              \textbf{74.86}   \\
  \end{tabular}}}
  \end{table*}

\section*{\textbf{Results on AIR-SARShip-1.0 dataset}}
\begin{figure*}[!h]
  \centering
  \includegraphics[width=1\textwidth]{"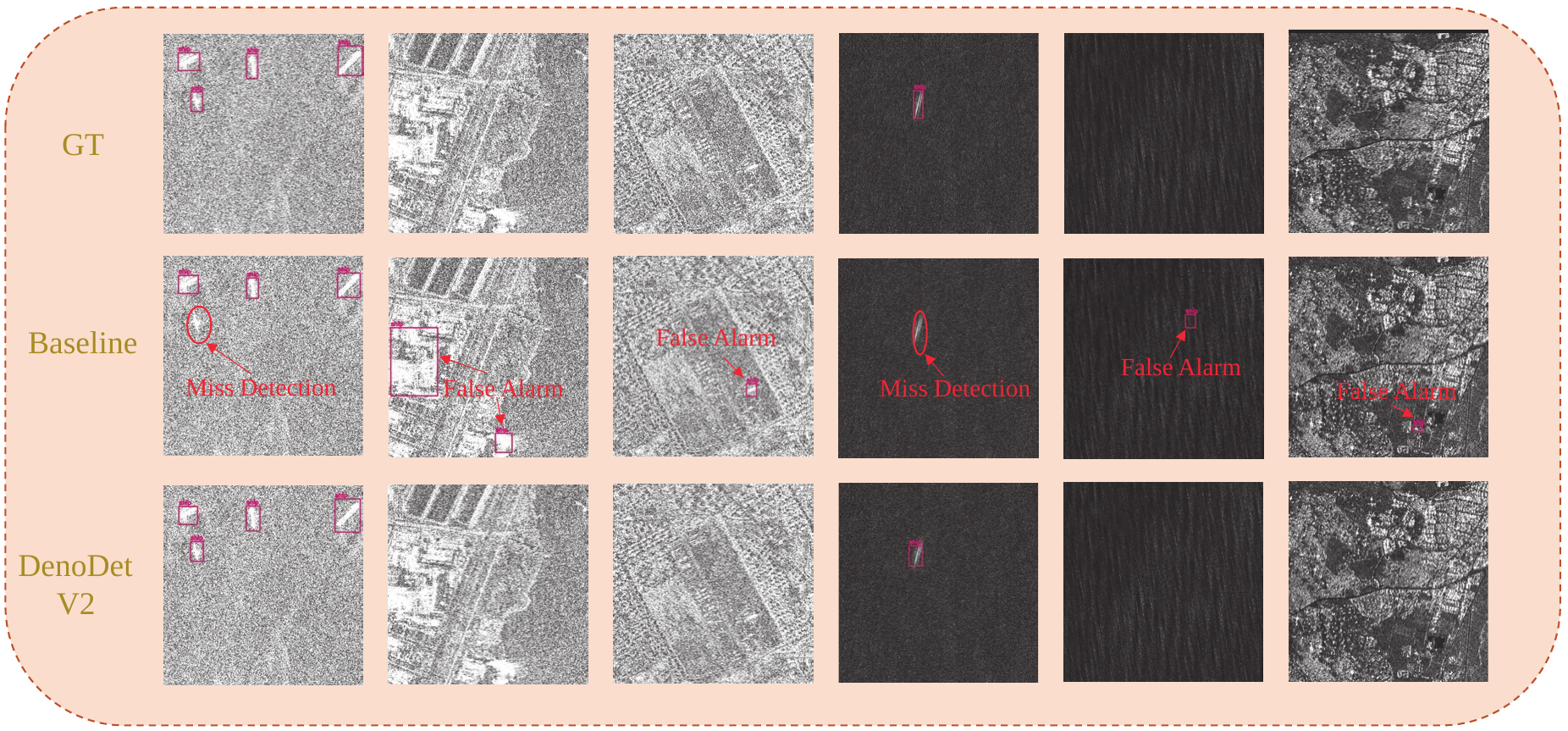"}
  \caption{Comparative visualization results 
  Qualitative comparison of miss detections between our DenoDet V2 and the baseline method on AIR-SARShip-1.0 dataset. Rectangular boxes represent predicted bounding boxes while circular annotations indicate miss detected instances. Our method significantly reduces miss detections across various challenging scenarios. This demonstrates DenoDet V2's enhanced capability to detect previously overlooked instances.
  }
  \label{fig:airsarship_sup}
\end{figure*}

We visualized the results on the AIR-SARShip-1.0 dataset, as shown in Fig. \ref{fig:airsarship_sup}. DenoDet V2 can effectively distinguish between background and object information in noisy areas, resulting in fewer missed detections and false alarms.

\clearpage

\section*{\textbf{Visualization Comparison}}

\begin{figure*}[h]
  \centering
    \vspace*{-1\baselineskip}
  \subfloat[DFT frequencies rank]{
    \includegraphics[width=0.5\textwidth]{
      "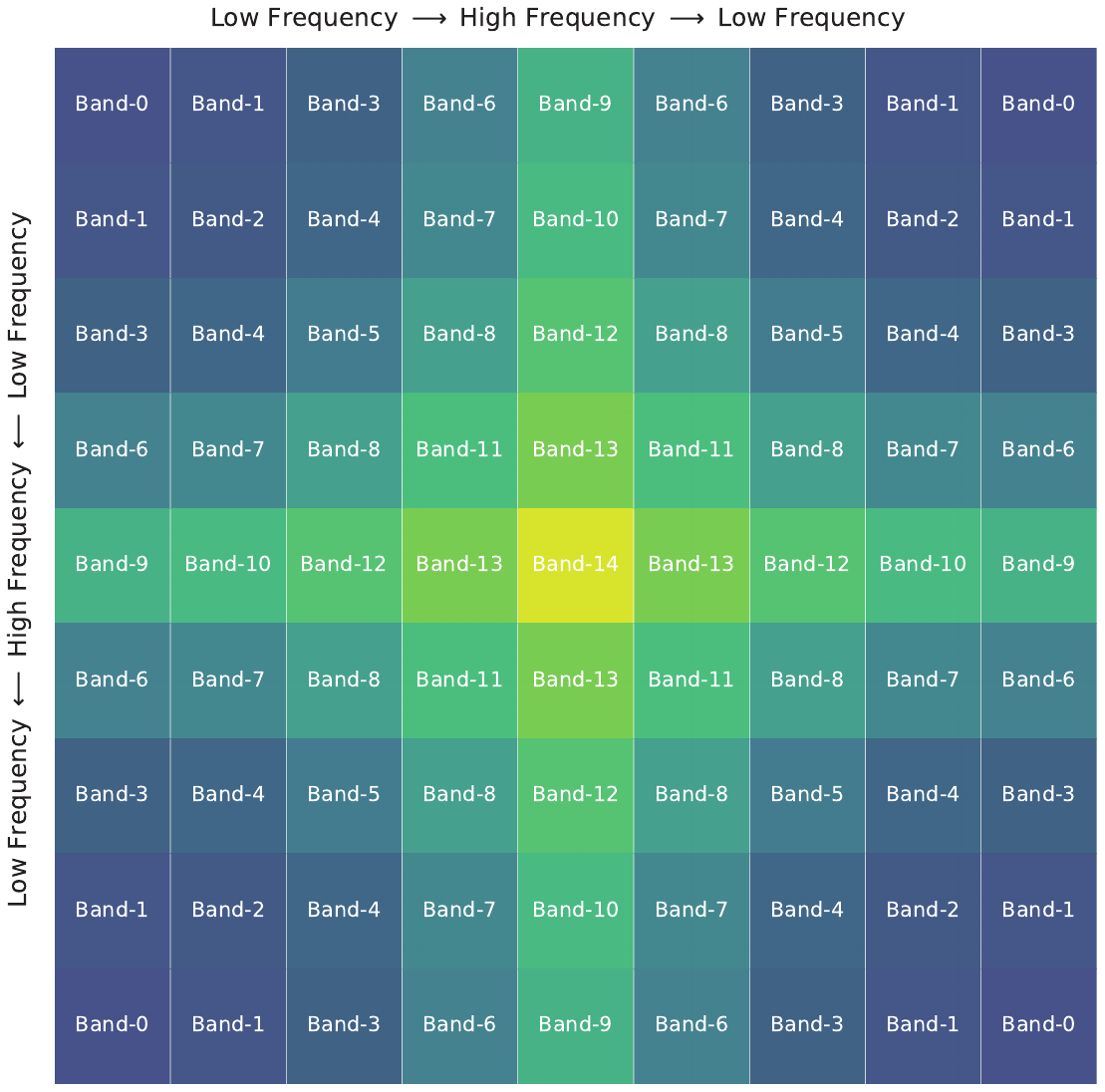"}
      \label{subfig:dft-filters}
  }
  \subfloat[After central shift]{
    \includegraphics[width=0.5\textwidth]{
      "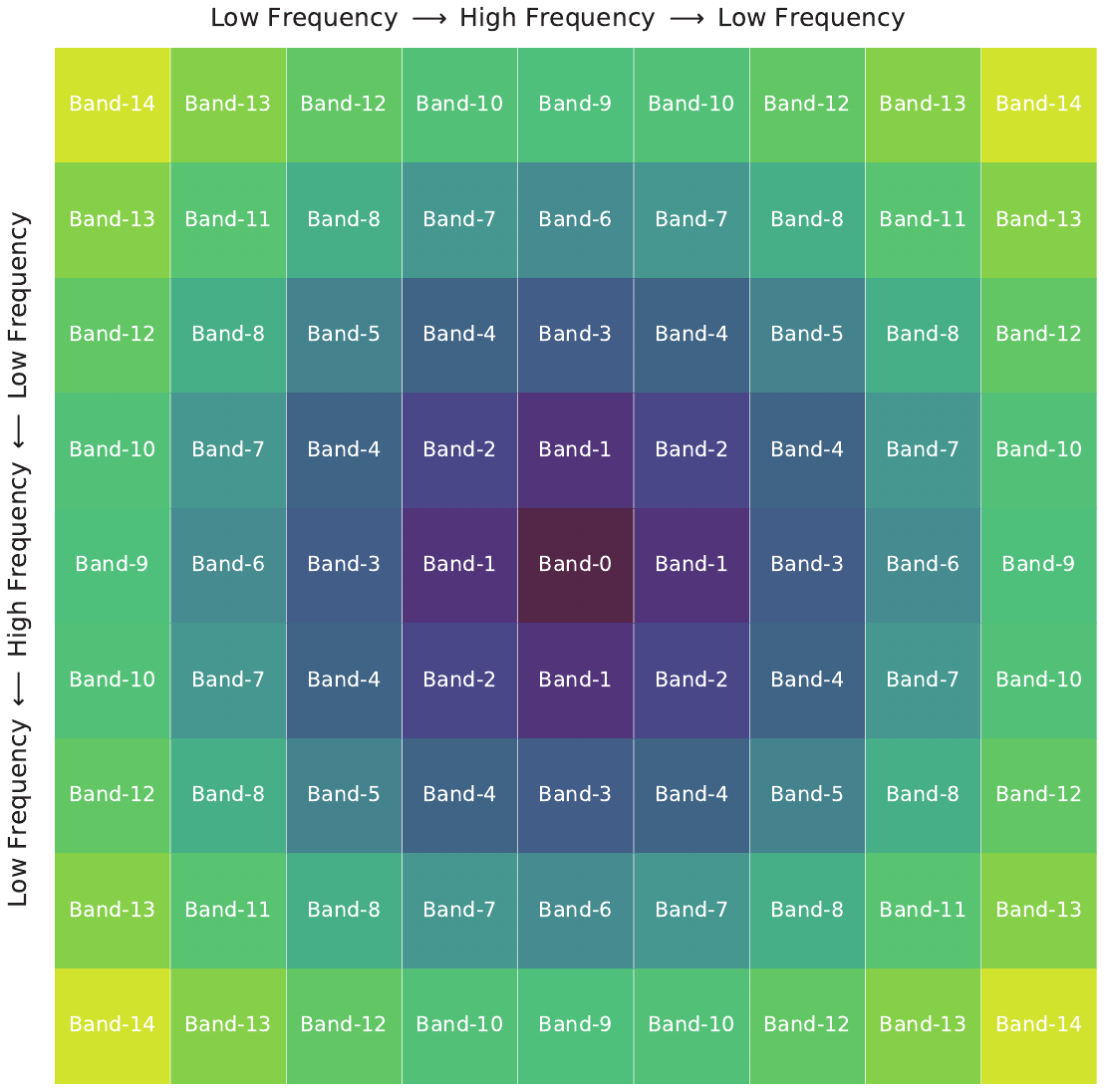"}
      \label{subfig:amplitude-bands}
  }
  \caption{Spectral decomposition of frequencies rank after DFT.
  (a) Illustration of native frequency distribution in the DFT-transformed domain. (b) Schematic representation of frequency spectrum distribution of the signal after zero-frequency centering.}
  \label{fig:dft}
  \vspace*{-1\baselineskip}
\end{figure*}

\end{document}